\documentclass{article}

\PassOptionsToPackage{sort, numbers, compress}{natbib}

\usepackage[final]{neurips_2024}

\usepackage[utf8]{inputenc} 
\usepackage[T1]{fontenc}    
\usepackage{hyperref}       
\usepackage{url}            
\usepackage{booktabs}       
\usepackage{amsfonts}       
\usepackage{nicefrac}       
\usepackage{microtype}      
\usepackage{xcolor}         
\usepackage{amsmath}
\usepackage{graphicx}
\usepackage{subfigure}
\usepackage{multirow}
\usepackage{booktabs}
\usepackage{makecell}
\usepackage{wrapfig}
\usepackage[normalem]{ulem}

\usepackage{float} 
\usepackage{xcolor}
\definecolor{hwj}{RGB}{67,151,143} 
\definecolor{zzw}{RGB}{135,206,235}

\usepackage{listings}
\newcommand{\model}{Extralonger}
\newcommand{\STRepresentation}{Unified Spatial-Temporal Representation}

\title{\model: Toward a Unified Perspective of Spatial-Temporal Factors for \\ Extra-Long-Term Traffic Forecasting}

\author{
Zhiwei Zhang$^{1}$, Shaojun E$^{1}$, \\
\textbf{Fandong Meng$^{2}$, Jie Zhou$^{2}$, Wenjuan Han$^{1}$\thanks{Corresponding author.}} \\
\\
$^{1}$Beijing Key Lab of Traffic Data Analysis and Mining, Beijing Jiaotong University, Beijing, China \\
$^{2}$WeChat AI, Tencent Inc., Beijing, China \\
\texttt{ \{zhiweizhang, 23140102, wjhan\}@bjtu.edu.cn,}
\\
\texttt{\{fandongmeng, withtomzhou\}@tencent.com}
}

\begin{document}
\maketitle

\begin{abstract}
     Traffic forecasting plays a key role in Intelligent Transportation Systems, and significant strides have been made in this field. However, most existing methods can only predict up to four hours in the future, which doesn't quite meet real-world demands. we identify that the prediction horizon is limited to a few hours mainly due to the separation of temporal and spatial factors, which results in high complexity. Drawing inspiration from Albert Einstein’s relativity theory, which suggests space and time are unified and inseparable, we introduce \model, which unifies temporal and spatial factors. \model~notably extends the prediction horizon to a week on real-world benchmarks, demonstrating superior efficiency in the training time, inference time, and memory usage. It sets new standards in long-term and extra-long-term scenarios. The code is available at \url{https://github.com/PlanckChang/Extralonger}.
\end{abstract}

\section{Introduction}

    Traffic forecasting stands as a pivotal endeavor within Intelligent Transportation Systems (ITS). This task aims to capture the spatial-temporal dynamics in traffic road networks by analyzing historical time steps across monitoring stations and subsequently forecasting future time steps.~\citep{workshopReviewerDemand1, workshopReviewerDemand2, STGNNs} Previous works denote raw traffic data as $\mathbf{X} \in \mathbb{R}^{T \times N \times C}$, where $T$ signifies the time steps length, $N$ represents the count of monitoring stations, and $C$ denotes the raw feature channel. They conceptualize traffic road networks as graphs, with nodes representing monitoring stations and edges between two nodes indicating connectivity. The early works have achieved significant progress in short-term traffic forecasting, whose prediction horizon is within one hour.

    In real-world ITS, there exists a considerable demand for prediction horizons that far exceed the typically short-term duration (usually $\le1$ hour) offered by prevailing deep learning methods. While SSTBAN~\citep{sstban} has initiated exploration into extended forecasting periods of 2 -- 4 hours, introducing a long-term traffic forecasting task, we propose an extra-long-term task with the prediction horizons ranging from 0.5 days to 1 week. In this real-world context, the traffic data incurs substantial computational and memory overheads due to the $T \times N$ feature set. Almost all previous approaches to fusing and processing this data exhibit a critical limitation due to treating temporal and spatial dimensions distinctly. Separate processing paradigms for temporal and spatial dimensions necessitate repetitive computations across dimensions, leading to inefficiencies. Specifically, temporal feature processing necessitates spatial iterations, while spatial feature processing requires temporal iterations. Consequently, the computational complexity is one order higher than the magnitude of pure time series prediction tasks, e.g., $\mathcal{O}(NT^{2}+TN^{2})$ in the self-attention mechanism, while memory consumption expands to $\mathcal{O}(TN^2 + NT^2)$, as shown in Table~\ref{tab:complexityComparision}. Such computational and memory demands pose significant challenges to extend the prediction horizon into the extra-long-term scenario.

     Fortunately, we draw inspiration from Albert Einstein's relativity theory~\citep{relativity} (See Appendix~\ref{appendix:inspiration} for more details), on how to make a \textbf{unified} perspective on spatial-temporal factors to decrease the computational complexity and memory consumption. We claim that spatial and temporal factors are unified and inseparable, and therefore, they should be treated simultaneously. To address this, we propose the Unified Spatial-Temporal Representation, which circumvents the need to expand dimensionality for embeddings, by incorporating the spatial feature directly into the same time step and the temporal feature into the corresponding node. Using this tailored representation, we design a three-route Transformer-based architecture for traffic forecasting, named \model. Our contributions are threefold:
    
    \begin{itemize}
        \item We propose \model, founded on Unified Spatial-Temporal Representation. Additionally, we introduce the extra-long-term traffic forecasting task.
        
        \item \model~achieves a one-order reduction in both computational complexity, $\mathcal{O}(NT^{2}+TN^{2}) \rightarrow \mathcal{O}({T^2+N^2})$, and memory usage, $\mathcal{O}(NT^2 + TN^2) \rightarrow \mathcal{O}(T^2+N^2)$. \model~tremendously excels the baselines in terms of memory consumption, training efficiency, and inference speed. Especially, in the longest step scenario, \model~achieves a 172$\times$ reduction in memory usage, 500$\times$ increase in training speed, 385$\times$ increase in inference speed than the prior best performance method (Figure~\ref{fig:costComparison}).
        
        \item \model~first successfully extends the traffic forecasting horizon from 2-4 hours to 1 week. It outperforms baselines in both long-term scenarios (2-4 hours, as shown in Table~\ref{longtermcomparison}) and extra-long-term scenarios (0.5 days to 1 week, detailed in Table~\ref{extralongcomparison}).
    \end{itemize} 
\begin{figure}[t]
  \centering
\includegraphics[width=1\textwidth]{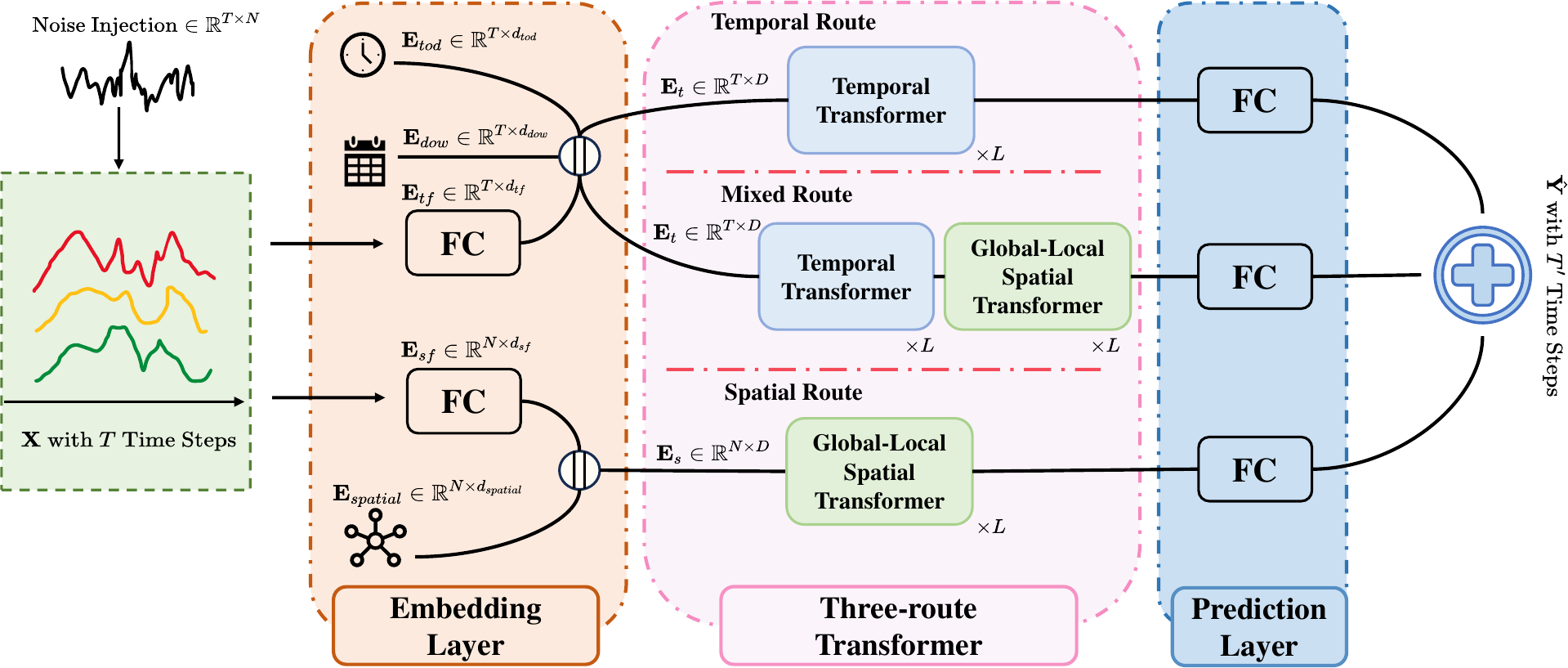}
  \caption{Overview of the architecture.}
  \label{overview}
  \vspace{-6mm}
\end{figure}
\section{Preliminary}

     Given traffic road network $\mathcal{G} = (\mathcal{V}, \mathcal{E})$, where $\mathcal{V}$ is the set of $N$ nodes (i.e., monitoring stations) and $\mathcal{E}$ is the set of edges. An edge exists if two nodes are connected, from which we derive the adjacency matrix $\mathcal{A}$. The historical traffic signal data are represented as $\mathbf{X} = [\mathbf{x}_0, \mathbf{x}_1, \ldots, \mathbf{x}_{T-1}] \in \mathbb{R}^{T\times N\times C}$, where $T$ denotes the length of historical time steps, $C$ represents the raw feature, with a value of 1 for a specific feature such as flow, speed or occupation. 

    The goal of traffic forecasting is to find a model $\mathbb{F}$ to predict the future traffic signal data $ \hat{\mathbf{Y}} = [ \hat{\mathbf{x}}_{T}, \hat{\mathbf{x}}_{T+1}, \ldots, \hat{\mathbf{x}}_{T+T'-1} ]$, where $T'$ is the length of future time steps, which can be formulated as:
    \begin{equation}
     \mathbf{X} = [\mathbf{x}_0, \mathbf{x}_1, \ldots, \mathbf{x}_{T-1}] \xrightarrow{\mathbb{F}(\cdot | \Theta)} \hat{\mathbf{Y}} = [ \hat{\mathbf{x}}_{T}, \hat{\mathbf{x}}_{T+1}, \ldots, \hat{\mathbf{x}}_{T+T'-1} ]
    \end{equation}
    where $\Theta$ is the parameters, conditional on which, the model $\mathbb{F}(\cdot)$ has the minimum loss calculated with the ground truth $\mathbf{Y}$.

\section{\model}\label{sec:model}
    We first introduce the Unified Spatial-Temporal Representation in Section \ref{sec:representation}, followed by a step-by-step description of the architecture of \model~in Section \ref{sec:three-route-transformer}.
\subsection{Unified Spatial-Temporal Representation}\label{sec:representation}
\subsubsection{Classical Representation}\label{sec:classicalComplexity}\label{sec:classicalrepresentation}
\begin{wrapfigure}{r}{0.45\textwidth}
    \vspace{-0.14in}
      \centering
      {\includegraphics[width=0.45\textwidth]{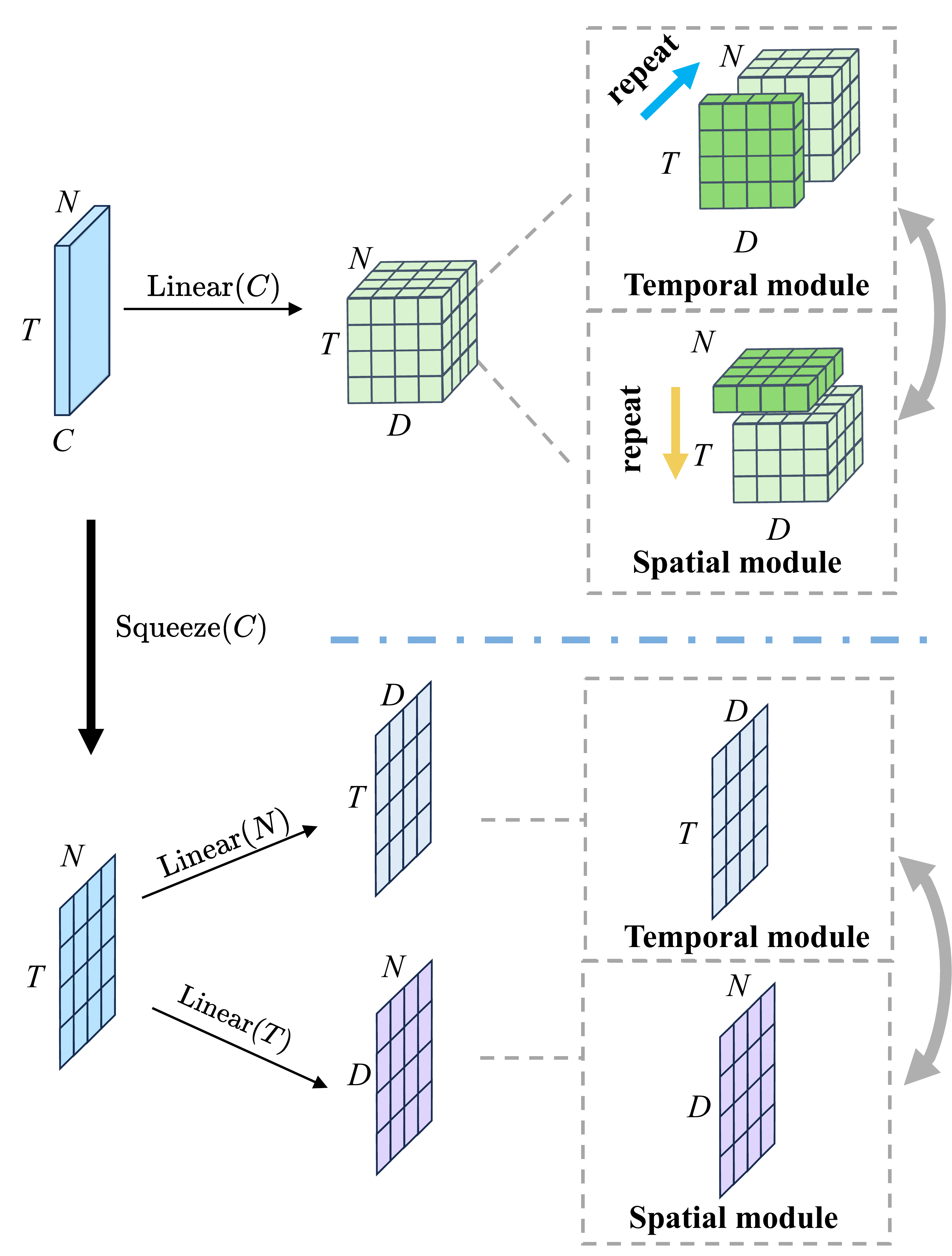}}
        \caption{Classical pipeline (TOP) and New pipeline (BOTTOM).}
      \label{fig:pipeline}
      \vspace{-0.08in}
    \end{wrapfigure}

    The pipeline with the classical representation $\mathbf{E} \in \mathbb{R}^{T \times N \times D}$ for previous traffic forecasting models is illustrated at the top of Figure~\ref{fig:pipeline} and formulated as follows :
    \begin{equation}
    \mathbf{X} \in \mathbb{R}^{T \times N \times C} \xrightarrow{\mathrm{Linear}(C)}\mathbf{E} \in \mathbb{R}^{T \times N \times D} \xrightarrow{\mathbb{F}(\cdot | \Theta  )} \hat{\mathbf{Y}} \in \mathbb{R}^{T'\times N\times C}
    \label{equ:classicalpipeline}
    \end{equation}
    where $D$ is the feature dimension. All current traffic forecasting methods, to the best of our knowledge, utilize the classical representation. They linearly project the raw data to the $\mathbf{E}$, which serves as the input to the forecasting model $\mathbb{F}$ involving temporal and spatial modules. The combination of the two modules is summarized as temporal-first, spatial-first, and spatial-temporal simultaneous~\cite{autost}, as depicted by the gray bidirectional arrow in Figure~\ref{fig:pipeline}.
    
    We categorize prior methods into four main groups: RNN-based methods, attention-based methods, Transformer-based methods, and CNN-based methods. The former three groups of methods necessitate spatial iterations when processing temporal features, and require temporal iterations when processing spatial features. Essentially, while processing features for one dimension, the other dimension acts as a role of batch size. Consequently, the computational complexity of RNN-based methods is $\mathcal{O}(TN)$. Considering the diverse variants of attention-based and Transformer-based methods, we only give the complexity of the representative self-attention mechanism, i.e., $\mathcal{O}(NT^2 + TN^2)$. 
    
    Memory usage analysis is complex and depends on many factors beyond the scope of our paper, including the parallelization framework, optimizer type, and decisions regarding intermediate state maintenance~\citep{memoryUsage}. Ignoring the constant memory footprint of model parameters and optimizer state, we approximate memory usage as $\mathcal{O}({TN})$ for RNN-based methods and $\mathcal{O}(NT^2 + TN^2)$ for attention-based and Transformer-based methods.
    
    Especially, CNN-based methods~\citep{graphwarvenet, dmstgcn} can treat the spatial and temporal dimensions jointly. Similar to image data $\mathbb{R}^{Height \times Width \times Channel}$ in computer vision, CNN-based methods correspondingly process the traffic data $\mathbb{R}^{T \times N \times D}$. Therefore, the computational complexity of CNN-based methods follows the complexity of image processing, resulting in $\mathcal{O}({k^2TN})$, where $k$ denotes the kernel size. Since CNN-based methods leverage GNNs as the spatial module, the total complexity is $\mathcal{O}(k^2TN + TN)$ for time complexity and $\mathcal{O}({TN})$ for memory usage. 
    
    These limitations of time and space complexity restrict the ability of existing methods to extend the prediction horizon to a weekly level. More details are given in Appendix~\ref{appendix:complexity} and Table~\ref{tab:complexityComparision}.

\subsubsection{New Representation}
    To address high-complexity issues, we rethink the classical representation and propose the Unified Spatial-Temporal Representation. Since feature number $C$ equals 1, we squeeze it directly. We adopt two fully connected linear layers and combine them with other prior embeddings (Section~\ref{sec:embedding}) to obtain the temporal and spatial representations: $\mathbf{E}_t$ and $\mathbf{E}_s$. With these two representations, the new pipeline is shown at the bottom of Figure~\ref{fig:pipeline} and formulated as follows:
    \begin{align}
      \mathbf{X}\in \mathbb{R}^{T\times N \times C}\xrightarrow{\mathrm{Squeeze}(C)} \mathbb{R}^{T\times N}  
    \left \{
    \begin{aligned}
    &  \xrightarrow{\mathrm{Linear}(N)}  \mathbf{E}_{t}  \in \mathbb{R}^{T \times D} \\
    &  \xrightarrow{\mathrm{Linear}(T)} \mathbf{E}_{s}  \in \mathbb{R}^{N \times D} \\
    \end{aligned}
    \right \}
     \xrightarrow{\mathbb{F}(\cdot |\Theta)} 
    \hat{\mathbf{Y}} \in \mathbb{R}^{T'\times N}
    \label{equ:newpipeline}
    \end{align}
     The new representation integrates the whole nodes to the same time step directly and integrates all the time steps to the same node as the linear layers play a channel-mixed role. Hence, for every $i$ in the range $[0, T-1]$, each element in the set $\{\mathbf{E}_{t}[i,j]: j\in[0,D-1]\}$ contains information about all the nodes at time step $i$. Similarly, for every $n$ in the range $[0, N-1]$, each element in the set $\{\mathbf{E}_s[n, m]: m\in[0,D-1]\}$ contains information about all the time steps at node $n$. This operation reflects the idea of \textbf{unification}: each time step incorporates information from all nodes, while each node integrates information across all time steps, which contrasts with the classical representation that relies solely on the feature itself at a particular node and time step. Early methods that treat spatial and temporal dimensions separately require repetitive operations, leading to increased complexity. However, the Unified Spatial-Temporal Representation overcomes this issue by reducing one dimension. Additionally, our novel representation offers the following three advantages.

    \textbf{Complexity Reduction.}\quad The Unified Spatial-Temporal Representation inherently reduces computational complexity and memory usage compared to the classical representation. This stems from that one dimension less than the classical representation. This eliminates the need for repetitive operations across the other dimension. Since our model employs a Transformer-based architecture denoted as $\mathbb{F}$, the time complexity decreases from $\mathcal{O}(NT^{2}+TN^{2})$ to $\mathcal{O}({T^2+N^2})$. Notably, if incorporating efficient attention mechanisms~\citep{efficencyTransformer}, the computational complexity could be further reduced to $\mathcal{O}({TlogT + NlogN})$. Similarly, memory usage reduces from $\mathcal{O}(NT^2 + TN^2)$ to $\mathcal{O}(T^2 + N^2)$.

    \begin{table}[t]
      \centering
        \caption{Comparison of time complexity, space complexity and maximum path lengths of mainstream methods. $T$ is the time steps length, $N$ is the node number and $k$ is the kernel size of CNN. The maximum path length of the self-attention mechanism with classical representation scales as $\mathcal{O}(2)$ due to independent computations along the temporal and spatial dimensions.}
      \label{tab:complexityComparision}
      \begin{tabular}{@{}lccc@{}}
      \toprule
      Method                                     & Time Complexity            & Space Complexity   & Maximum Path Length              \\ \midrule
      RNN-based (Classical)                                        & $\mathcal{O}(TN)$     & $\mathcal{O}(TN)$  & $\mathcal{O}(T+N)$         \\
      CNN-based (Classical)                                   & $\mathcal{O}(k^2TN+TN)$       & $\mathcal{O}(TN)$   & $\mathcal{O}(log_kT+log_kN)$ \\
      Self-Attention (Classical)                   & $\mathcal{O}(NT^2 + TN^2)$ & $\mathcal{O}(NT^2 + TN^2)$    & $\mathcal{O}(2)$               \\
      Self-Attention (Unified)                & $\mathcal{O}(T^2+N^2)$     & $\mathcal{O}(T^2 + N^2)$   & $\mathcal{O}(1)$               \\ \bottomrule
      \end{tabular}
      \vspace{-5mm}
    \end{table}
    
\begin{figure}[t]
      \centering
      \subfigure[The aggregation comparison.]{
        \includegraphics[width=.45\textwidth]{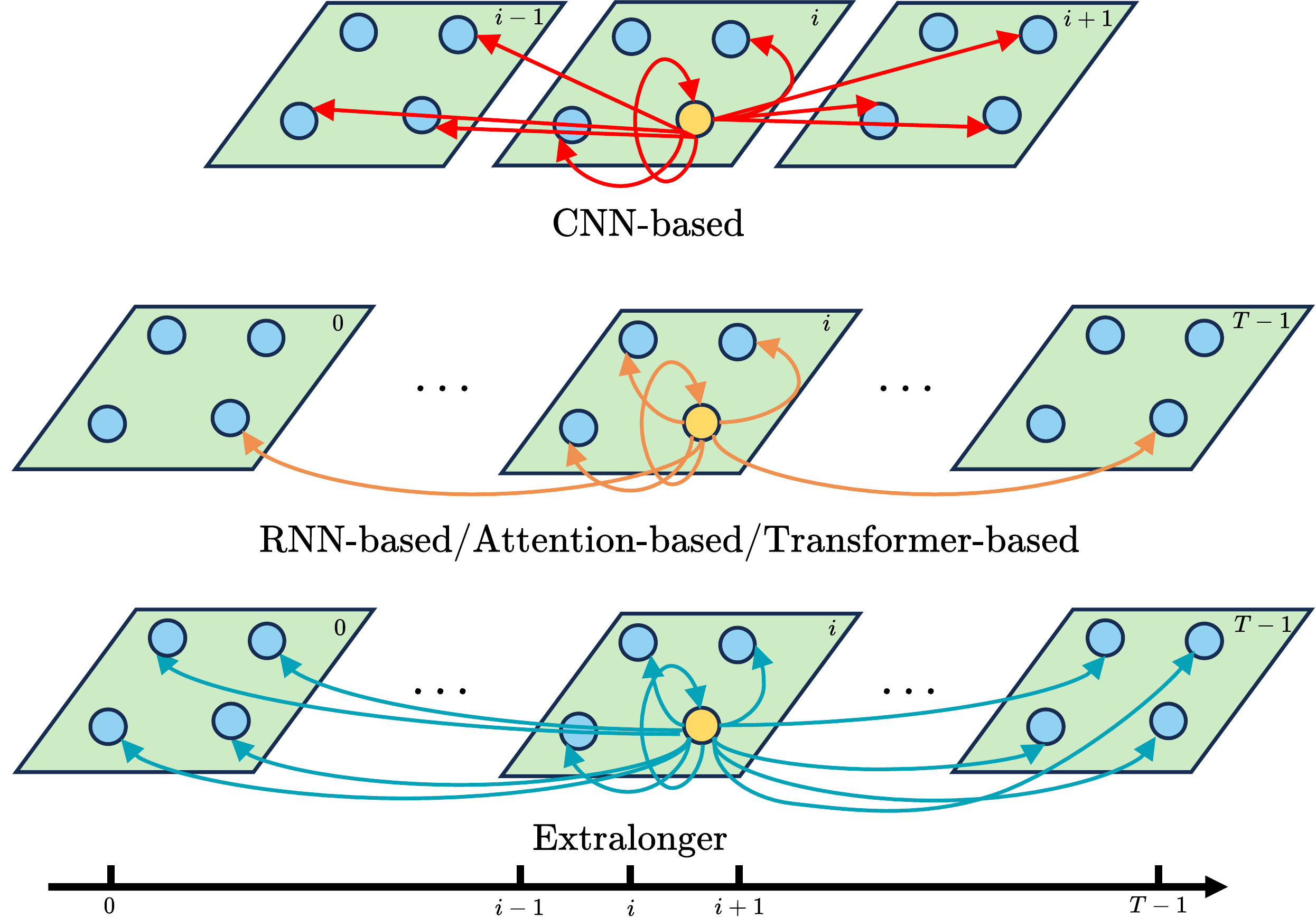}
        \label{fig:aggregationComparison}
      }  
      \qquad
      \subfigure[The receptive field comparison.]{
        \includegraphics[width=.4\textwidth]{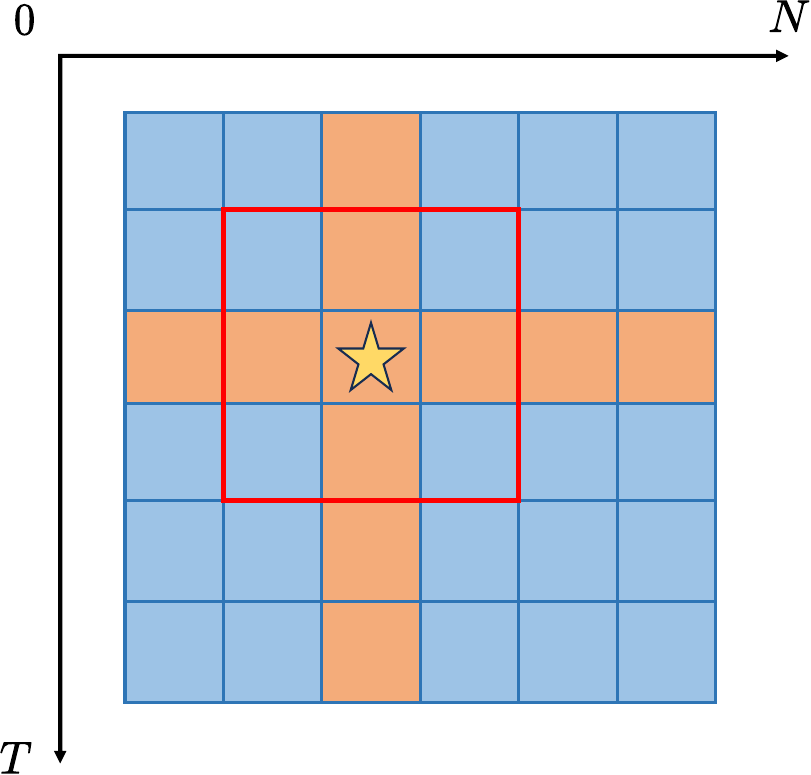}
        \label{fig:receptiveField}
      }
      \caption{Given the yellow circle and star are the analysis target at $t$ time step, (a) only \model~aggregates along different nodes and different time steps globally; (b) CNN-based methods aggregate the information within the restricted receptive field (red box), and the other early methods aggregate along the temporal and spatial dimensions separately (two orange belts). Capturing temporal and spatial dependency globally with the \STRepresentation~(whole area), \model~does simultaneously and efficiently.}
      \vspace{-4mm}
    \end{figure}

    \textbf{Simultaneous Aggregation.}\quad The traffic signal transmits between the upstream nodes and the downstream nodes with time going by, so the signal is related to different nodes and different times, not just fixing one dimension and varying another one. \model~allows each node to be simultaneously linked to all other nodes at the current time step, as well as to the nodes from all historical and future time steps. This representation enables \model~to effectively capture the complex spatial-temporal dynamics underlying traffic trends. Conversely, RNN-based, attention-based and Transformer-based methods can only aggregate the nodes at the same time step or the same node among different time steps. Treating traffic data as a grid, CNN-based methods are unable to simultaneously consider all nodes at the same time step. Instead, they can only aggregate the nodes within adjacent time steps due to the inherent receptive field. The aggregation comparison of ours and other methods is shown in Figure~\ref{fig:aggregationComparison}.

    \textbf{Complete Receptive Field.}\quad
    The importance of a complete receptive field from both spatial and temporal dimensions for capturing long-range dependencies in traffic forecasting has been well-established in prior work~\cite{dmstgcn, pdformer, autost}. \model~achieves a complete receptive field compared to existing approaches. This stems from the inherent characteristic of the Unified Spatial-Temporal Representation and the self-attention mechanism employed by \model. As illustrated in the three-route Transformer architecture in Section~\ref{sec:three-route-transformer}, the self-attention mechanism enables comprehensive connectivity among all elements in the sets $\{\mathbf{E}_{t}[i,:]: i \in [0,T-1]\}$ and $\{\mathbf{E}_{s}[n,:]: n \in [0,N-1]\}$. This full connection facilitates the aggregation of information from any node across the entire road network among all time steps, i.e., the complete receptive field, as depicted by all the squares in Figure~\ref{fig:receptiveField}. Consequently, the maximum operation path length in \model~is $\mathcal{O}(1)$. In contrast, RNN-based, attention-based and Transformer-based methods typically aggregate information along the temporal and spatial dimensions separately, leading to an axial receptive field (orange belts in Figure~\ref{fig:receptiveField}). CNN-based methods are constrained by their inherent limitations on convolution size (red box in Figure~\ref{fig:receptiveField}). While CNN-based methods can operate along both dimensions, they cannot make sure all nodes in the complete road network and all time steps are in the receptive field. Moreover, unlike the image size that is fixed or can be preprocessed to the same size in computer vision tasks, CNN-based models are not robust enough due to the $T$ varying with the prediction horizon. 
  
\subsection{Architecture of \model}\label{sec:three-route-transformer}

    Founded on the \STRepresentation, we propose \model, comprising three parts: embedding layer, three-route Transformer, and prediction layer, as shown in Figure~\ref{overview}. 
    The embedding layer transforms the raw traffic data into an embedding space. The resulting embedding consists of two components, the spatial representation $\mathbf{E}_s$ and the temporal representation $\mathbf{E}_t$. The three-route Transformer is the heart of \model~and is comprised of three parallel routes: a temporal route, a spatial route, and a mixed route. Notably, all three routes can simultaneously process information from both spatial and temporal dimensions due to the adoption of the \STRepresentation. Each route emphasizes specific aspects of the data. The temporal Transformer in the temporal route and the mixed route is implemented as a standard Transformer encoder, and its detailed description is omitted for brevity. The Global-Local Spatial Transformer is specifically designed for the spatial route and the mixed route. Finally, projected by the prediction layer first, the outputs of the three routes are added up with a hand-crafted weight.

\subsubsection{Embedding Layer}\label{sec:embedding}
    We first inject noise into $\mathbf{X}$ with learnable parameters (More details in Appendix~\ref{appendix:noiseInjection}). Then, we use two fully connected linear layers to transform $\mathbf{X}$ to feature embeddings: $\mathbf{E}_{tf}\in\mathbb{R}^{T\times d_{tf}}$ and $\mathbf{E}_{sf}\in \mathbb{R}^{N \times d_{sf}}$. We also utilize learnable embedding to capture the periodicity. Specifically, we employ $\mathbf{E}_{tod}\in \mathbb{R}^{T \times {d_{tod}}}$ for the 24-hour cycle (namely, timestamp-of-day embedding) and $\mathbf{E}_{dow}\in \mathbb{R}^{T\times d_{dow}}$ for the 7-day cycle (namely, day-of-week embedding). We adopt a learnable spatial embedding $\mathbf{E}_{spatial}\in \mathbb{R}^{N\times d_{spatial}}$, introduced by GWNet~\citep{graphwarvenet}, to capture the spatial dynamics. The temporal representation $\mathbf{E}_t$ is formed by concatenating the aforementioned embeddings: $\mathbf{E}_t = \mathbf{E}_{tf} || \mathbf{E}_{tod} || \mathbf{E}_{dow} \in \mathbb{R}^{T \times D}$, where $||$ denotes concatenation. The spatial representation $\mathbf{E}_s$ is informed by concatenating embeddings: $\mathbf{E}_s = \mathbf{E}_{sf} || \mathbf{E}_{spatial} \in \mathbb{R}^{N \times D}$.
\subsubsection{Global-Local Spatial Transformer}\label{sec:globallocalatt}
    \begin{wrapfigure}{r}{0.5\textwidth}
    \vspace{-0.45in}
      \centering
      {\includegraphics[width=0.45\textwidth]{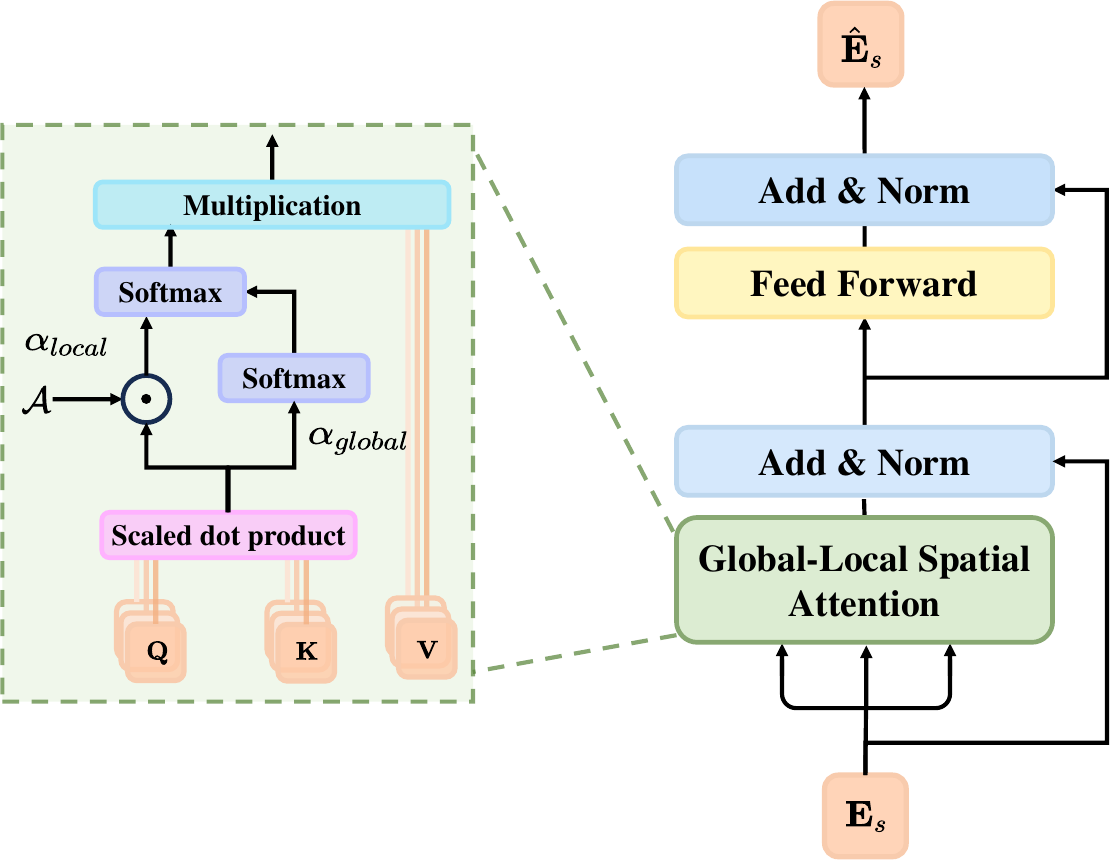}}
      \caption{Global-Local Spatial Transformer.}
      \label{globallocalatt}
      \vspace{-0.12in}
    \end{wrapfigure}

    The proposed Global-Local Spatial Transformer is specifically tailored to exploit the inherent topological characteristics of the spatial domain. It integrates the road network's prior information into the self-attention mechanism. By incorporating dynamic dependencies derived from all nodes and focusing on neighboring nodes to aggregate local correlations, the Global-Local Spatial Transformer explicitly leverages both global and local spatial information, as shown in Figure~\ref{globallocalatt}. More details about motivation and insights are given in Appendix~\ref{appendix:globallocal}.

    Given $\mathbf{Q} = \mathbf{E}_s\mathbf{W}_{Q}$, $\mathbf{K} = \mathbf{E}_s\mathbf{W}_K$, $\mathbf{V} = \mathbf{E}_s\mathbf{W}_V$, where $\mathbf{W}_Q$, $\mathbf{W}_K$ and $\mathbf{W}_V$ are learnable parameters, the Global-Attention score before softmax is followed as:
    \begin{equation}
      \alpha_{global}(\mathbf{Q}, \mathbf{K}) = \frac{\mathbf{Q}\mathbf{K}^T}{\sqrt{D}}
    \end{equation}
    Then, we inject the road network's prior information by using the adjacency matrix $\mathcal{A}$ as a mask to obtain the Local-Attention score before softmax, formulated as:
    \begin{equation}
      \alpha_{local} = \alpha_{global} \odot \mathcal{A}
    \end{equation}
    where $\odot$ denotes the element-wise product. We designate it as \textit{local} because only information from adjacent nodes is considered, while non-adjacent nodes are masked out. Both $\alpha_{local}$ and $\alpha_{global}$ are subjected to a softmax function to obtain normalized attention weights. Global-Local Spatial Attention (GLSAtt) is obtained by weighted summation with $\mathbf{V}$.
    \begin{equation}
      \mathrm{GLSAtt} = (\mathrm{Softmax}(\alpha_{local}) + \mathrm{Softmax}(\alpha_{global}))\frac{\mathbf{V}}{2}
    \end{equation}
    We sequentially apply layer normalization (LN), skip connections, and a feed-forward network (FFN), the same procedure as the vanilla Transformer, to yield the output $\hat{\mathbf{E}}_s$.
    \begin{align}
     \mathbf{Z} & = \mathrm{LN}(\text{GLSAtt}(\mathbf{E}_s)) + \mathbf{E}_s \\ 
    \hat{\mathbf{E}}_s & = \mathrm{LN}(\mathrm{FFN}(\mathbf{Z})) + \mathbf{Z}
    \end{align}
    
\subsubsection{Prediction Layer}
    We employ linear layers to project the output of the three-route Transformer to predict the corresponding $T'$ time steps, which is necessary for the asymmetrical $T-T'$ setup. Subsequently, the projected outputs are aggregated using hand-crafted weights, resulting in $\hat{\mathbf{Y}}\in\mathbb{R}^{T'\times N}$. The Huber loss~\citep{huber} is then calculated between $\hat{\mathbf{Y}}$ and the ground truth $\mathbf{Y}$, formulated as:
    \begin{equation}
    \text{HuberLoss} = 
    \begin{cases}
     \frac{1}{2}(\mathbf{Y} - \hat{\mathbf{Y}})^2,                   & \text{if } \left | \mathbf{Y} - \hat{\mathbf{Y}} \right | \le \delta \\
    \delta \cdot \left( \left | \mathbf{Y} - \hat{\mathbf{Y}} \right | - \frac{1}{2}\delta\right), & \text{otherwise}
    \end{cases}
    \end{equation}
    where $\delta$ is a positive value.
    
\section{Experiments}
\subsection{Datasets and Experiment Setup}
We evaluate the performance of our proposed \model~in long-term and extra-long-term traffic forecasting scenarios using three real-world datasets: PEMS04, PEMS08 and Seattle Loop, which serve as widely adopted benchmarks for traffic prediction tasks. Following early works, we split each dataset into training, validation, and testing sets using a ratio of 6:2:2. \footnote{Because of the page limitation, the experiment results of Seattle Loop are in Appendix~\ref{appendix:seattle}.}
Following SSTBAN~\citep{sstban}, in long-term scenarios, both $T$ and $T'$ are set to 24/36/48. In the extra-long-term scenarios, $T$ and $T'$ are set to 144/288/576/864/1152/1440/1728/2016. Full details are given in Appendix~\ref{appendix:datasetsAndSetup}. We selected three widely used metrics: root mean square error (RMSE), mean absolute error (MAE), and mean absolute percentage error (MAPE).

\subsection{Baselines}
We carefully choose the following baselines in long-term scenarios.
(1) \textbf{HA}: Historical Average, which predicts the future traffic flow by averaging the historical data.
(2) \textbf{VAR}~\citep{var}: Vector Auto-Regression, which is a classical multivariate time series forecasting method.
(3) \textbf{DCRNN}~\citep{dcrnn}: Diffusion Convolutional Recurrent Neural Network, predicting with diffusion convolutions and GRU.
(4) \textbf{GWNet}~\citep{graphwarvenet}: Graph WaveNet, combining the dilated casual convolutions and graph convolutions to capture dynamics correlation.
(5) \textbf{GMAN}~\citep{gman}: Graph Multi-Attention Network, especially using spatial attention with randomly partitioned vertices group.
(6) \textbf{AGCRN}~\citep{agcrn}: Adaptive Graph Convolutional Recurrent Network, which learns node-specific patterns and avoids the pre-defined graphs.
(7) \textbf{DMSTGCN}~\citep{dmstgcn}: Dynamic and Multi-faceted Spatial-Temporal Graph Convolution Network, which captures the dependency with a spatial learning method and multi-faceted fusion module enhancement.
(8) \textbf{SSTBAN}~\citep{sstban}: Self-Supervised Spatial-Temporal Bottleneck Attentive Network, utilizing a bottleneck attention scheme to reduce the computational cost.

In extra-long-term scenarios, because the device could not afford the resource cost of the deep learning baselines, we focus on comparing the performance of \model~with established statistical methods: HA and VAR. 
\subsection{Model Implementation and Training Details}\label{sec:implementation}
\model~utilizes the Adam optimizer~\citep{adam} with the default learning rate of 0.0001, which would decay with predefined milestones. The batch size is set to 16. The hyperparameters remain consistent for different scenarios on the same benchmark, and the layer number $L$ for per route is set to 1. We adopt Huber loss~\citep{huber} as the loss function, which leverages the strengths of MAE and MSE with $\delta=1$ while addressing their limitations. The outputs from the three routes (temporal, spatial, and mixed) are combined using a weighted sum, with weights of 0.25, 0.25, and 0.50, respectively. \model~is implemented in PyTorch and all experiments were conducted on one single NVIDIA 2080Ti GPU. 

\begin{table}[t]
  \centering
  \caption{Performance comparison in long-term scenarios. The smaller the value, the better.}
  \label{longtermcomparison}
  \resizebox{\textwidth}{!}{
  \begin{tabular}{@{}cc|ccc|ccc|ccc@{}}
  \toprule
  \multirow{2}{*}{Dataset} &  \multirow{2}{*}{Mehtod}  & \multicolumn{3}{c|}{24 Time Steps}    & \multicolumn{3}{c|}{36 Time Steps}    & \multicolumn{3}{c}{48 Time Steps}     \\  
                           &       & \textit{RMSE}  & \textit{MAE}   & \textit{MAPE}      & \textit{RMSE}  & \textit{MAE}   & \textit{MAPE}      & \textit{RMSE}   & \textit{MAE}   & \textit{MAPE}      \\      \midrule
  \multirow{8}{*}{PEMS04}  & HA          & 81.57 & 56.47 & 45.49     & 106.58& 76.01 & 68.84     & 127.28 & 93.37 & 94.62     \\
                           & VAR         & 41.09 & 27.19 & 21.42     & 45.44 & 30.48 & 24.51     & 49.46  & 33.5  & 27.28     \\
                           & DCRNN       & 42.86 & 28.70 & 21.23     & 51.40 & 33.78 & 27.10     & 57.85  & 38.26 & 33.73     \\
                           & GWNet       & 35.52 & 22.79 & 16.04     & 38.17 & 24.71 & 17.67     & 40.60  & 26.42 & 18.99     \\
                           & GMAN        & 38.10 & 21.67 & 17.78     & 52.86 & 22.12 & 16.43     & 47.85  & 23.35 & 17.98     \\
                           & AGCRN       & 34.44 & 21.63 & 14.65     & 38.19 & 24.15 & 16.33     & 38.26  & 24.18 & 16.31     \\
                           & DMSTGCN     & 32.09 & 20.32 & 14.13     & 34.86 & 22.47 & 15.86     & 35.05  & 22.50 & 16.56     \\
                           & SSTBAN      & 32.82 & 20.17 & 14.43     & 34.15 & 20.82 & 14.83     & 35.51  & 21.66 & 15.90     \\
                           & \textbf{\model} & \textbf{31.89} & \textbf{19.60} & \textbf{13.60}     & \textbf{32.96} & 
                           \textbf{20.31} & \textbf{14.12}     & \textbf{34.01}  & \textbf{20.97} & \textbf{14.42}     \\ \midrule

  \multirow{8}{*}{PEMS08}  & HA          & 69.72 & 48.3  & 32.09     & 92.72 & 65.99 & 46.64     & 111.85 & 81.51 & 61.29     \\
                           & VAR         & 44.47 & 28.31 & 19.53     & 48.96 & 31.7  & 22.56     & 52.14  & 34.51 & 25.28     \\
                           & DCRNN       & 33.34 & 22.60 & 15.46     & 39.37 & 25.82 & 18.53     & 45.64  & 30.47 & 25.10     \\
                           & GWNet       & 29.47 & 19.07 & 12.25     & 33.54 & 21.76 & 13.68     & 34.20  & 22.60 & 14.16     \\
                           & GMAN        & 34.29 & 17.38 & 15.66     & 35.89 & 17.21 & 16.33     & 48.54  & 18.70 & 16.81     \\
                           & AGCRN       & 28.05 & 17.45 & 11.25     & 30.96 & 19.39 & 12.73     & 31.11  & 19.46 & 12.88     \\
                           & DMSTGCN     & 26.55 & 16.75 & 11.44     & 28.50 & 18.15 & 12.64     & 28.94  & 18.34 & 12.93     \\
                           & SSTBAN      & 26.32 & 15.97 & 12.29     & 28.30 & 16.84 & 12.20     & 28.82  & \textbf{16.94} & 12.47     \\
                           & \textbf{\model} & \textbf{26.29} & \textbf{15.86} & \textbf{10.40}     & \textbf{27.64} & \textbf{16.57} & \textbf{11.34}     & \textbf{28.77}  & 17.14 & \textbf{11.54}     \\ \bottomrule
  \end{tabular}}
  \vspace{-6mm}
  \end{table}

\subsection{Performance Comparison}\label{sec:performance}
 The performance results are presented in Table~\ref{longtermcomparison} and Table~\ref{extralongcomparison} for long-term and extra-long-term scenarios, respectively. The best results are shown in bold. 

Our proposed \model~outperforms baselines across almost all long-term scenarios. \model~consistently surpasses SSTBAN, the second best model, on the PEMS04 dataset, exhibiting an average reduction of 2.36\% in RMSE, 2.82\% in MAE, and 5.95\% in MAPE. Similarly, on PEMS08, \model~outperforms SSTBAN by an average margin of 0.87\% in RMSE, 0.37\% in MAE, and 7.35\% in MAPE. The only exception is the MAE in the PEMS08-48 scenario, where \model~achieves the second-best result, marginally behind SSTBAN.

In the extra-long-term scenarios, \model~demonstrates significant superiority over HA and VAR across all three evaluation metrics. This wide margin in performance shows the effectiveness of \model~in capturing extra-long-term temporal dependencies and spatial relationships. In PEMS04, our model delivers steady results when the input time step $T$ is fixed at 288 (a whole day), whereas in PEMS08, the errors rise gradually. This can be attributed to the smaller node count in PEMS08, which enables the model to achieve better performance in smaller-step scenarios.

An interesting observation in extra-long-term scenarios is that, unlike \model~obtaining the best performance at $T'=144$ (length of half a day), the performance of HA and VAR exhibits a decline at $T'=144$ compared to longer steps. We posit that this decline arises from a potential misalignment between the input and output data. They struggle to predict future trends when the input data contains significantly inverse trends from the predicted future steps. 
Take a special sample as an example, whose input is 144 steps from the first half of the day and the output is the traffic flow in the second half of the day. It is difficult for HA and VAR to predict the second half solely on the first half.

\begin{table}[t]
  \centering
  \caption{Performance comparison in extra-long-term scenarios. TS: Length of Time Steps.}
  \label{extralongcomparison}
  \resizebox{\linewidth}{!}{
  \begin{tabular}{@{}cc|ccc|ccc|ccc@{}}
  \toprule
  \multirow{2}{*}{Dataset} &  \multirow{2}{*}{TS}  & \multicolumn{3}{c|}{HA}       & \multicolumn{3}{c|}{VAR}  & \multicolumn{3}{c}{\textbf{\model}} \\
                           &  & \textit{RMSE}   & \textit{MAE}    & \textit{MAPE}     & \textit{RMSE}  & \textit{MAE}   & \textit{MAPE}     & \textit{RMSE}     & \textit{MAE}     & \textit{MAPE}       \\ \midrule
  \multirow{8}{*}{PEMS04}  & 144      & 177.11 & 144.67 & 220.60   & 77.22 & 53.63 & 64.87    & \textbf{39.33}    & \textbf{23.66}  & \textbf{16.04}      \\
                           & 288      & 128.50 & 103.95 & 166.55   & 50.43 & 34.98 & 34.22    & \textbf{40.46}    & \textbf{24.18}  & \textbf{16.70}      \\
                           & 576      & 129.35 & 104.44 & 166.73   & 52.90 & 36.99 & 35.55    & \textbf{42.15}    & \textbf{26.08}   & \textbf{19.32}      \\
                           & 864      & 129.65 & 104.56 & 167.80   & 54.55 & 38.32 & 37.43    & \textbf{42.98}   & \textbf{26.60} & \textbf{19.36}      \\
                           & 1152     & 129.78 & 104.56 & 169.59   & 55.99 & 39.50 & 39.12    & \textbf{42.98}    & \textbf{26.88}   & \textbf{19.33}      \\
                           & 1440     & 129.91 & 104.66 & 169.81   & 56.64 & 40.05 & 39.40    & \textbf{42.30}   &\textbf{26.62}   &\textbf{19.24}      \\
                           & 1728     & 130.00 & 104.77 & 169.29   & 57.01 & 40.40 & 39.69    &\textbf{43.33}  & \textbf{27.09}   & \textbf{19.63}      \\
                           & 2016     & 130.37 & 105.09 & 167.48   & 56.21 & 39.64 & 39.23    & \textbf{43.26}    & \textbf{27.52}   &\textbf{20.21}      \\ \midrule

  \multirow{8}{*}{PEMS08}  & 144      & 153.98 & 123.81 & 114.18   & 89.00 & 61.09 & 46.52    &\textbf{32.17}    &\textbf{18.84}   &\textbf{12.85}      \\
                           & 288      & 112.80 & 89.66  & 87.33    & 56.07 & 38.51 & 26.06    &\textbf{32.76}    &\textbf{19.38}  &\textbf{13.05}      \\
                           & 576      & 113.80 & 90.29  & 87.28    & 63.16 & 43.37 & 28.59    &\textbf{37.73}    &\textbf{22.06}   &\textbf{14.23}      \\
                           & 864      & 114.36 & 90.47  & 87.09    & 65.07 & 45.25 & 30.91    &\textbf{38.57}    &\textbf{22.57}   &\textbf{15.04}      \\
                           & 1152     & 114.66 & 90.52  & 87.67    & 66.06 & 45.91 & 32.42    &\textbf{40.45}    &\textbf{23.50}  &\textbf{16.03}      \\
                           & 1440     & 114.97 & 90.80  & 88.71    & 67.96 & 47.03 & 32.69    &\textbf{41.44}    &\textbf{23.78}   &\textbf{17.90}      \\
                           & 1728     & 115.39 & 91.33  & 89.66    & 67.36 & 46.53 & 32.34    &\textbf{40.85}    &\textbf{23.68}  &\textbf{18.42}      \\
                           & 2016     & 115.43 & 91.54  & 90.01    & 65.45 & 44.95 & 31.37    &\textbf{41.38}    &\textbf{23.69}   &\textbf{19.18}      \\ \bottomrule
  \end{tabular}
  }
    \vspace{-6mm}
  \end{table}
\subsection{Resource Consumption Comparison}\label{sec:resourceComparison}

\begin{figure}[t]
  \centering
  \subfigure{
    {\includegraphics[width=.31\textwidth]{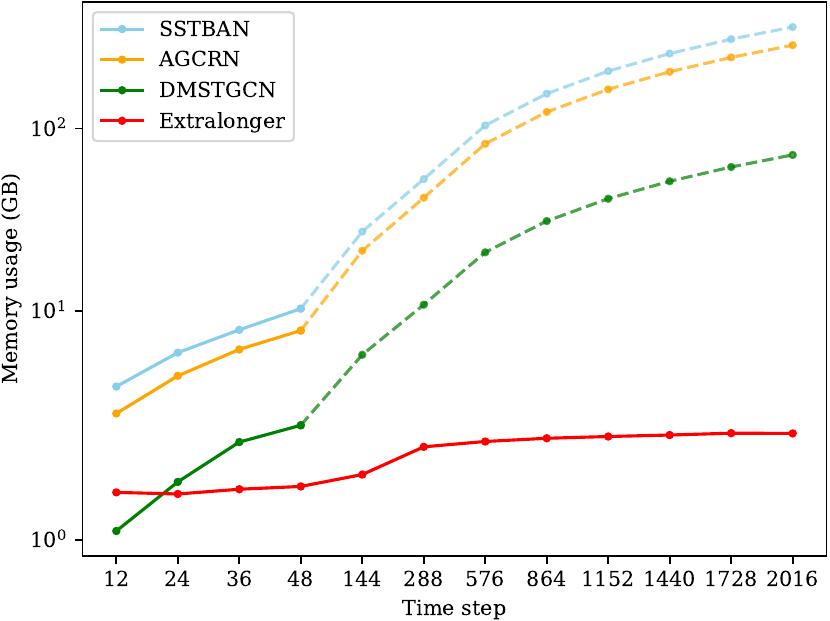}
    \label{memoryusage}}
  }  
  \subfigure{
    {\includegraphics[width=.31\textwidth]{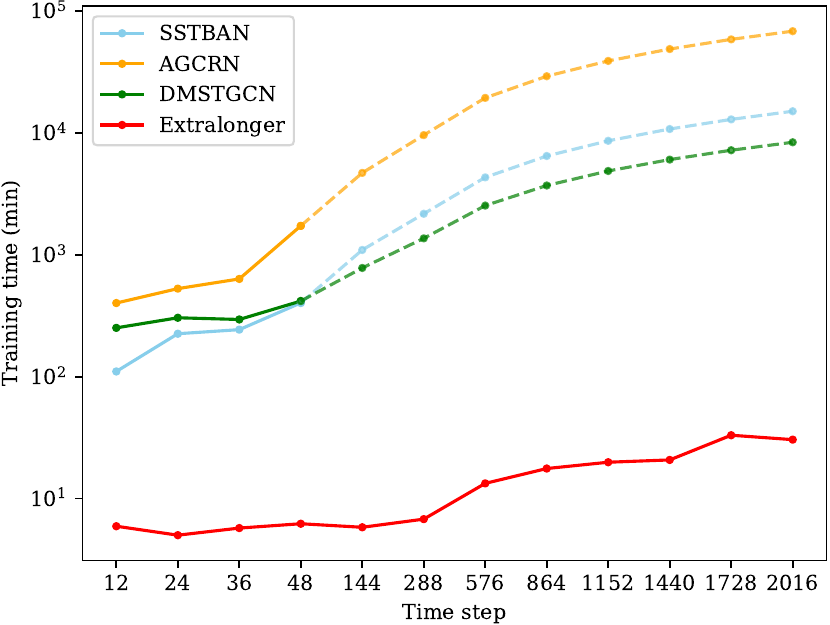}
    \label{trainingtime}}
  }
  \subfigure{
    {\includegraphics[width=.31\textwidth]{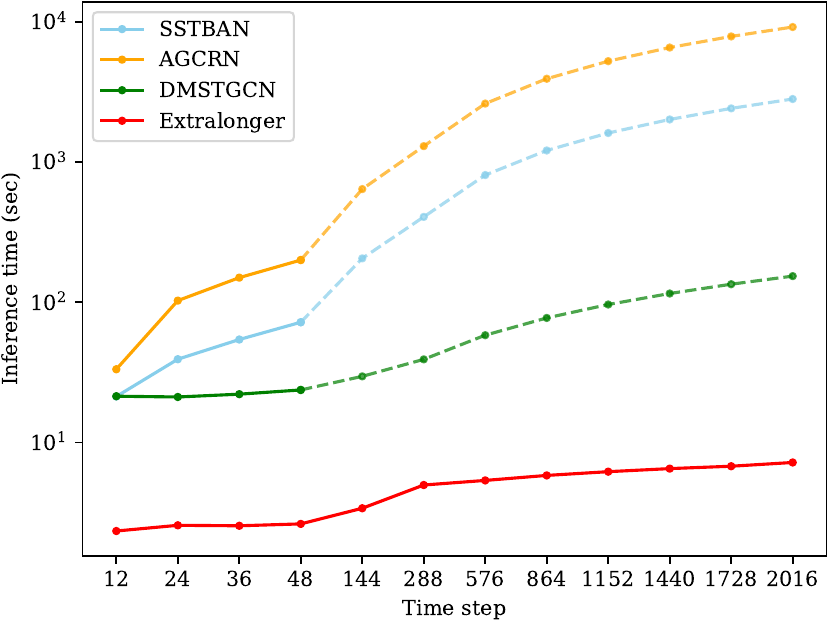}
    \label{inferencetime}}
  }
    \caption{Resource cost comparison w.r.t. memory usage (LEFT), training time (MIDDLE) and inference time (RIGHT). The y-axis represents the logarithm of the value. The dashed line reveals the fitted trendline based on actual results.}
  \label{fig:costComparison}
\end{figure}

Our proposed \model~demonstrates superior resource efficiency compared to existing methods, as illustrated in Figure~\ref{fig:costComparison}. The resource consumption comparison is conducted on the PEMS04 dataset. Due to computational limitations and the high cost of the baseline models, we report actual values (solid lines) for 12, 24, 36, and 48-step predictions. Leveraging the complexity analysis from Section~\ref{sec:classicalComplexity}, we extrapolate the resource consumption trends for extra-longer-term scenarios and show the trends using dashed lines in Figure~\ref{fig:costComparison}.

In long-term scenarios, \model~consumes on average 75.87\% less memory, 97.13\% less training time, and 93.53\% less inference time than SSTBAN across 12, 24, 36, and 48-step predictions. The sole exception is memory usage in the 12-step scenario, where the CNN-based method DMSTGCN achieves a marginally lower cost. We speculate that this is because parameter memory usage dominates at the shortest horizon, and DMSTGCN utilizes fewer parameters than our model. However, this advantage diminishes for longer prediction steps (>24 steps), where \model's memory efficiency becomes increasingly pronounced.

\begin{table}[H]
            \vspace{-.5mm}
		\caption{Ablation study.}
		\label{ablation}
		\centering
    \begin{tabular}{@{}l|ccc@{}}
      \toprule
      Method            & RMSE     & MAE      & MAPE     \\ \midrule
      w/o local \& global   & 37.12    & 22.69    & 15.97    \\
      w/o global        & 34.24      & 21.04    & 14.45    \\
      w/o local         & 34.82      & 21.31    & 14.93    \\
      w/o spatial route & 37.31   & 22.72    & 15.89    \\
      w/o temporal route & 34.85  & 21.31    & 14.30    \\
      w/o mixed route    & 35.59  & 21.53    & 14.44    \\
      w/o noise injection & 34.12 & 21.03 & 14.49    \\
      \textbf{\model} & \textbf{34.01}    & \textbf{20.97}    & \textbf{14.42}    \\ \bottomrule
    \end{tabular}%
  \vspace{-5mm}
\end{table}

In extra-long-term scenarios, \model~demonstrates its dominance in resource efficiency. In the 2016-step scenario, our model achieves remarkably low resource consumption, requiring only 2.1 GB of memory, 30.5 minutes for training, and 7.21 seconds for inference. With conservative estimates using fitted polynomial functions, our model's memory usage, training time, and inference time remain only 0.58\%, 0.20\%, and 0.26\% of SSTBAN's cost, respectively. These results strongly support the efficiency gains from the Unified Spatial-Temporal Representation employed by \model. 

\subsection{Ablation Study}\label{sec:ablation}

 We conducted an ablation study on PEMS04 in the 48-step scenarios. We remove the local part, global part, and global\&local part from Global-Local Spatial Transformer, in Table~\ref{ablation}. Our findings indicate that both the local and global modules play an important role. Moreover, we explore the necessity of each route in the three-route Transformer. The results show the importance of each route. The ablation of noise injection presents its effectiveness as well.

\section{Conclusion}

    We propose \model, a novel traffic forecasting model that leverages a unified perspective of spatial and temporal factors to address the limitations of current methods in resource efficiency. \model~achieves state-of-the-art performance in both long-term and extra-long-term scenarios. Moreover, it demonstrates significant reductions in resource consumption compared to previous models. Specifically, in the longest prediction horizon, memory usage, training time, and inference time are only 0.58\%, 0.20\%, and 0.26\% of the cost of the prior best method, respectively. Furthermore, \model~successfully extends the prediction horizon by a remarkable factor of 42 times (from 48 to 2016 steps). We believe that \model~establishes a new paradigm for traffic forecasting tasks and paves the way for the development of more efficient models for other spatial-temporal tasks.

\section{Acknowledgment}
The work described in this paper has been supported by the National Nature Science Foundation of China (No. 62406020).

\bibliography{neurips_2024}
\bibliographystyle{unsrtnat}

\clearpage
\appendix

\section{Related Work}

In early works, traffic forecasting is treated as a multivariate time series prediction task, so HA, ARIMA~\citep{arima}, and VAR~\citep{var} are the intuitive methods to model it, which neglect the inherent spatial dependencies within traffic data.

With the growing recognition of the equal importance of both spatial and temporal dimensions, traffic forecasting has evolved to consider road networks as graphs where node values (namely, flow, speed or occupancy) change over time. This paradigm shift led to the development of Spatial-Temporal Graph Neural Networks (STGNNs)~\citep{STGNNs} that leverage GNNs to capture spatial dependencies. STGNNs can be further categorized based on their approach to handling temporal information. RNN-based methods utilize Recurrent Neural Network or its variations to capture temporal dependencies~\citep{dcrnn, agcrn, stmetanet}. Attention-based methods~\citep{gman, ts-at, pdformer, astgcn, astgnn, autost, geoman} incorporate attention mechanisms to process temporal information. CNN-based methods~\citep{stgcn, graphwarvenet, stsgcn, dmstgcn} can handle spatial and temporal information simultaneously. However, one limitation of CNNs is their restricted receptive window, which can potentially hinder the processing of all nodes across extended temporal horizons. Beyond the aforementioned categories, some studies~\citep{stgode, STGNCDE} have focused on spatial information modeling. Additionally, efforts have been made to integrate differential equations with GNNs for enhanced spatial-temporal representation~\citep{dgcrn, ddSTGNN, dstagnn}. These efforts highlight the ongoing exploration of novel techniques for capturing the intricate interplay between spatial and temporal dynamics in traffic forecasting.

More recently, Transformer-based methods adopted the Transformer architecture to replace GNNs. Vanilla Transformer~\citep{transformer} is introduced for machine translation tasks. Recent years' works have focused on the promising prospect of Transformer-based models~\citep{graphtransformer,recipeforgraph} to replace the GNNs~\citep{gcn, gat, gat2, chebnet} for overcoming the over-smooth issue, which facilitates the evolution of the manner for spatial dimension processing in traffic forecasting field. Variants of Transformer begin to shine in the traffic forecasting field~\citep{sstban, pdformer, staeformer, st-transformer}.

Most related to ours is the work of AutoST~\citep{autost}, STSGCN~\citep{stsgcn} and SSTBAN~\citep{sstban}. AutoST and STSGCN notice the line of study that handles time and space separately. The former proposes a universal modeling framework to summarize the aggregation order issue. The latter constructs a new adjacency matrix with pre-defined steps, which limits the model's capability. SSTBAN is the first deep-learning-based work to research long-term traffic forecasting and uses a bottleneck attention scheme to reduce the computational cost to $\mathcal{O}({TN})$, but it has quadratic complexity on memory usage as well. 

Our work points out the deficiency in the classical representation of spatial-temporal systems, which results in high resource costs. We propose the Unified Spatial-Temporal Representation to unify the spatial and temporal dimensions to reduce the time and space complexity without any architectural modification, and successfully extend the prediction horizon to one week, following in the footsteps of SSTBAN.

\section{Inspiration from Relativity}\label{appendix:inspiration}
The Unified Spatial-Temporal Representation is inspired by the theory of relativity, a theory in physics that explains the universal, general relationship between space and time. Similarly, traffic forecasting inherently deals with uncovering the intertwined nature of spatial and temporal dimensions in traffic data. This realization motivated us to explore seemingly disparate fields to find solutions, mirroring the pursuit of knowledge across disciplines that foster scientific progress. 

Space and time are unified and inseparable in the relative view. In other words, when referring to one particular time step, we should also consider the spatial information, and vice versa. This translates directly to the context of traffic forecasting: understanding a specific time step necessitates simultaneous consideration of the corresponding spatial information, and vice versa. As the famous verb of Hermann Minkowski~\citep{minkowsk}, ``space for itself, and time for itself shall completely reduce to a mere shadow, and only some sort of \textbf{union} of the two shall preserve independence''. This implies that we should address both time and space concurrently (Unified Spatial-Temporal Representation), rather than isolating them (classical representation).

Thus, we propose the Unified Spatial-Temporal Representation to unify the spatial and temporal dimensions. As an additional explanation of the main body analysis, the $D$ in temporal representation $\mathbf{E}_t$ has spatial information, and the $D$ in spatial representation $\mathbf{E}_s$ has temporal information, which is the essence of the Unified Spatial-Temporal Representation. In this sense, we use the spatial information to define the time step and the temporal information to define the spatial node. From a data representation perspective, the data representation is 2D with Unified Spatial-Temporal Representation and 3D with classical representation.

\section{Complexity Analysis}\label{appendix:complexity}
The high-complexity issue in classical representation (as depicted in Equation~\ref{equ:classicalpipeline} and Figure~\ref{fig:pipeline}) hinders the prediction horizon extension. In the main body of this paper, we have concisely discussed the time and space complexity. We will give the full analysis in this section. We omit the batch size $B$. Focusing on the difference between the classical representation and the Unified Spatial-Temporal Representation, we focus on the core architecture with one single layer rather than delving into the details of each compared model. This setting allows for a more controlled comparison, highlighting the specific contribution of the proposed representation to the overall efficiency of the model.

\textbf{Time Complexity}\quad
We first analyze methods using the classical representation. For RNN-based methods, due to the repetitive operation for the spatial and temporal dimensions, the time complexity increases to $\mathcal{O}(TN)$. 

For attention-based methods, for example, Transformer architecture, the time complexity grows by one order to $\mathcal{O}(NT^2 + TN^2)$. In attention-based and Transformer-based methods, to simplify the analysis of various derivative models, we primarily concentrate on the basic self-attention mechanism for temporal and spatial dimensions, whose time complexity grows by one order to $\mathcal{O}(NT^2 + TN^2)$. The time complexity of CNN-based methods is influenced by the feature map size, kernel size and the number of channels. In our setting, the channel size corresponds to $D$. For fairness, we omit this part and only consider the feature map size and kernel size. We use the biggest feature map size, i.e., $T \times N$. Therefore, the time complexity is $\mathcal{O}(k^2TN)$, where $k$ is the kernel size. While the complexity of GNN is linear, the total complexity is $\mathcal{O}(k^2TN + TN)$.

When using the Unified Spatial-Temporal Representation, because we adopt the Transformer architecture in~\model, the time complexity is $\mathcal{O}(T^2+N^2)$ for the self-attention mechanism. 

\textbf{Space Complexity} 
As mentioned in the main body, we need to consider many factors in space complexity, such as the parallel framework, optimizer type, and whether to maintain the intermediate states, which is out of our scope. Therefore, we simplify it. Besides, we omit the parameter memory usage $M_{model}$, because we focus on the impact of data instead of the parameter memory usage for extra-long-term scenarios.

The space complexity of RNN-based methods and self-attention mechanisms scales linearly with the sequence length $T$ and the number of nodes $N$ due to the repetitive computations involved. These repetitive operations effectively act like a batch size and exhibit an approximately linear correlation with memory consumption. Consequently, the space complexity for RNN-based with classic representation is $\mathcal{O}(TN)$, while for attention-based and Transformer-based methods is $\mathcal{O}(NT^2 + TN^2)$.

For CNN-based methods, the original space complexity (ignoring the parameter memory usage) is typically $\mathcal{O}(Height \times Width)$, which can be high depending on the input data size. In the context of traffic forecasting, the space complexity is $\mathcal{O}(TN)$ as well.

With the Unified Spatial-Temporal Representation, the space complexity is $\mathcal{O}(T^2+N^2)$ in \model, which is a one-order reduction compared with the classical representation.   

\section{Why is Global-Local Spatial Transformer}\label{appendix:globallocal}
\textbf{Global Part}\quad
The GNN-based methods capture spatial dependency using static prior topology information, hypothesizing that the traffic flow trends among adjacent nodes are similar. However, we observe that non-adjacent nodes, such as Node 95 and 111, exhibit similar flow patterns. This suggests that dependencies can exist between nodes that are not spatial neighbors. To capture these long-range dependencies, \model~leverages a Transformer architecture with full connectivity between all nodes. With the Transformer, we treat the traffic road network as a fully connected graph, thus the $\alpha_{global}$ is gained. This global design enables \model~to go beyond the limitations of local neighborhoods and directly model relationships between any two nodes in the road network, regardless of their spatial proximity. The belief in learning global correlations is also supported by studies such as~\citet{sstban, graphwarvenet, gman}. 

\textbf{Local Part}\quad
While the proposed model captures long-range dependencies through the fully connected Transformer, the road network topology, represented by the adjacency matrix, remains a valuable source of prior knowledge. This matrix encodes the inherent spatial relationships between nodes, providing information about the correlations in traffic flow between adjacent nodes.  In essence, the model leverages the adjacency matrix to complement the spatial dependency modeling capabilities of the Transformer. Specifically, the adjacency matrix is injected as a mask to obtain the $\alpha_{local}$ due to they inherently have the same dimensionality. The local design draws inspiration from the works of~\citet{graphtransformer} and~\citet{recipeforgraph}. Refer to the Section~\ref{sec:globallocalatt} for other procedures.

\section{Learnable Noise Injection}\label{appendix:noiseInjection}
Caltrans Performance Measurement System~\citep{datasetCollection} and SSTBAN~\citep{sstban} both claim the importance of mitigating the impact of noise on the traffic forecasting model's performance. They acknowledge the inherent challenge of data validity in traffic forecasting. Data augmentation technique is a possible way to improve robustness. However, traditional data augmentation techniques commonly used in computer vision (CV)~\citep{cvAugmentation}, such as cropping, scaling, and rotation, are not well-suited for traffic data due to their inherent sequential nature and strong self-correlations. Additionally, complex deep learning methods like VAEs~\citep{vae} and GANs~\citep{gan} might be overly elaborate for this specific task.

To mitigate the impact of noise, we propose a simple yet effective data augmentation strategy -- involving the injection of learnable noise into the input data, which is implemented by the learnable parameter initialized with Xavier uniform distribution. We use this learnable noise embedding to imitate the impulse noise and enhance the model's robustness against real-world noise patterns. The result of the ablation experiment validates the effectiveness of this technique (Section~\ref{sec:ablation}).

\section{Details of Datasets and Setup}\label{appendix:datasetsAndSetup}
\begin{figure}[htp]
  \centering
  \subfigure[PEMS04]{
    {\includegraphics[width=.45\textwidth]{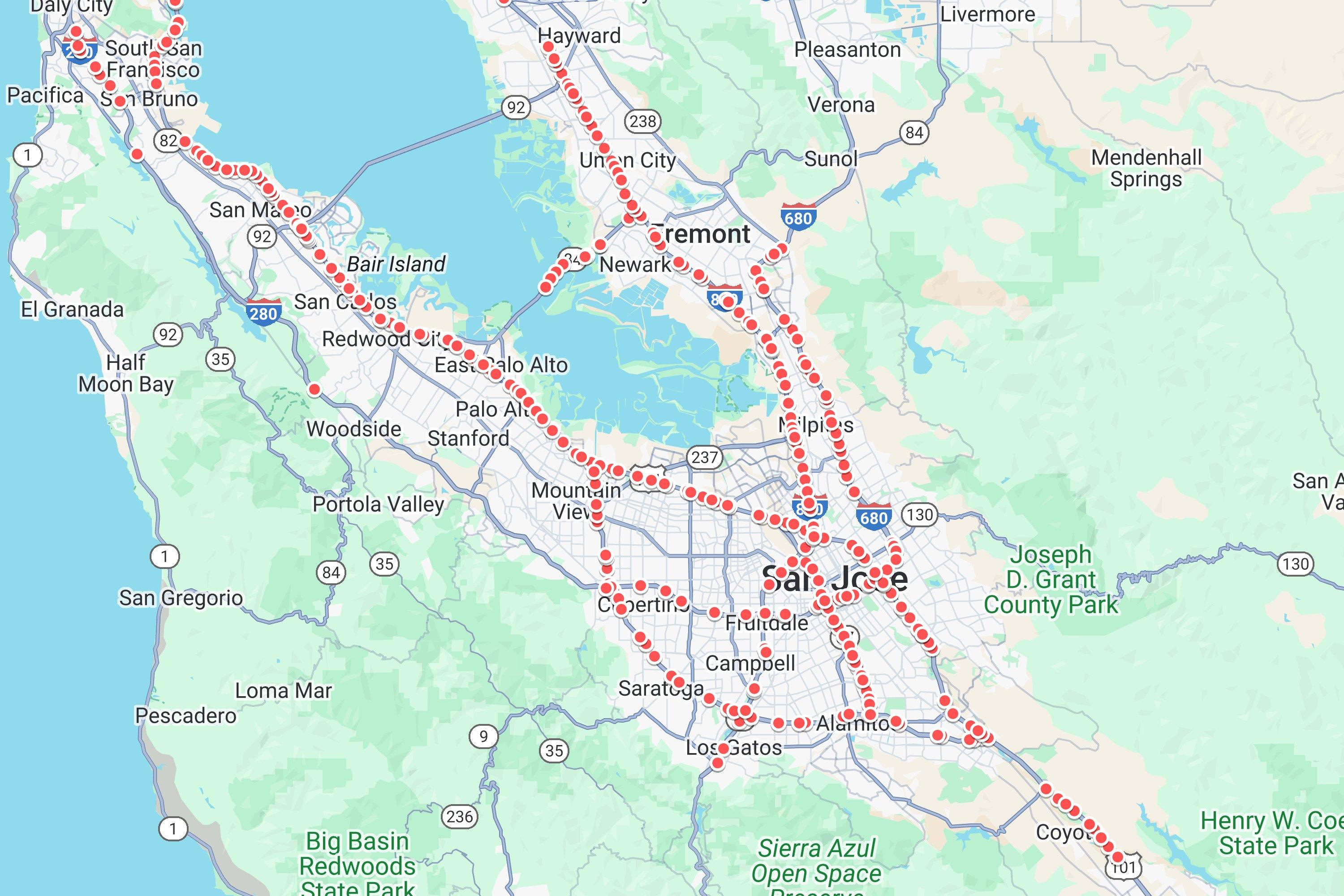}}
    \label{pems04map}
  }  
  \subfigure[PEMS08]{
    {\includegraphics[width=.45\textwidth]{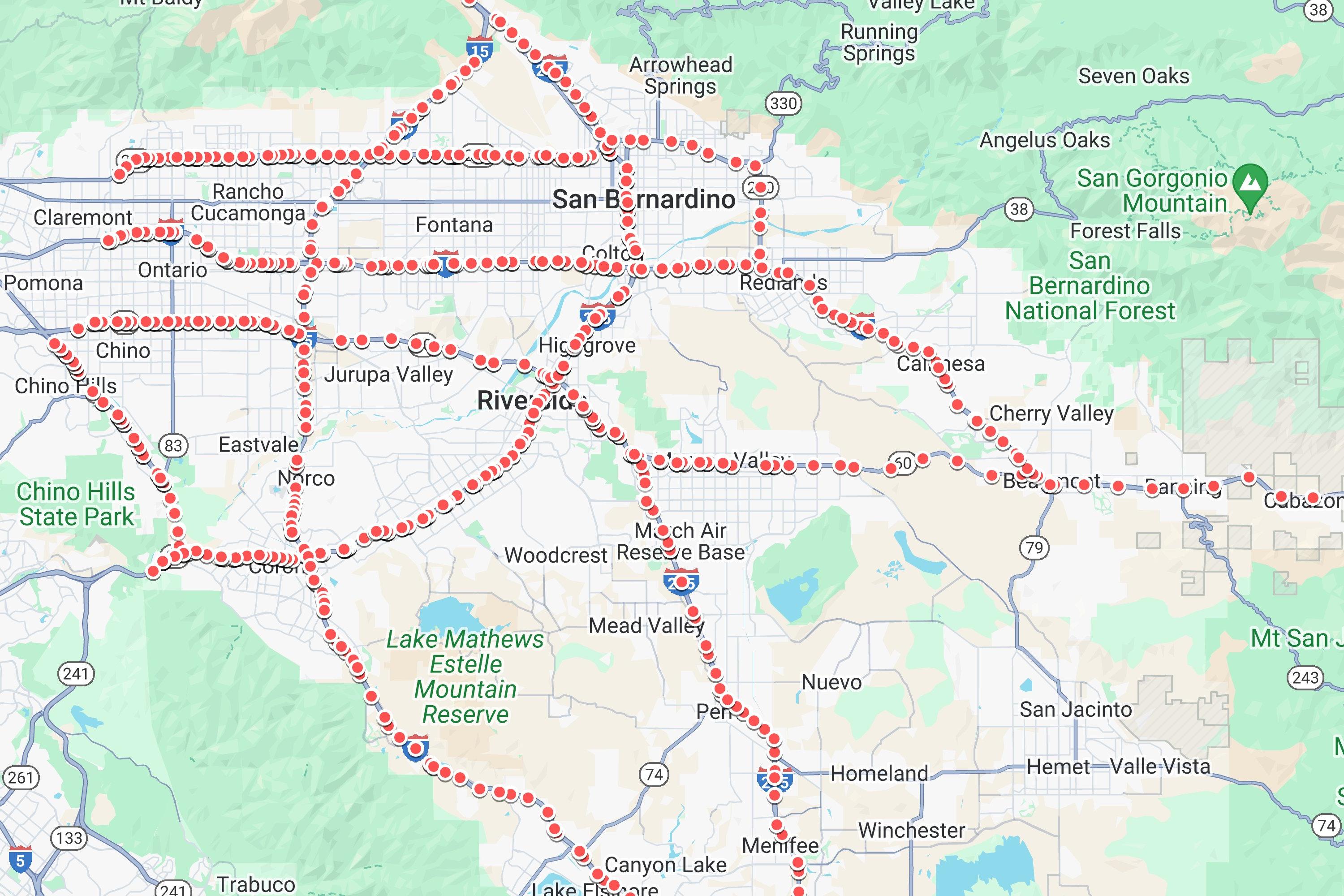}
    \label{pems08map}}
  }
  \caption{Node position of PEMS04 and PEMS08.}
  \label{map}
\end{figure}

Both PEMS04 and PEMS08 datasets are collected from the Caltrans Performance Measurement System~\citep{datasetCollection} at a 5-minute sampling frequency. Seattle Loop dataset is collected by the inductive loop detectors deployed on freeways in the Seattle area~\citep{seattle}. PEMS04 contains 307 nodes and 16992 time steps, covering the time range from 2018/01 to 2018/02. PEMS08 contains 170 nodes and 17856 time steps, covering the time range from 2016/07 to 2016/08. The Seattle Loop dataset consists of 323 nodes and 8,760 time steps, ranging from 2015/01 to 2015/12. To ensure a fair comparison, the setup of Seattle Loop follows SSTBAN~\citep{sstban} that aggregates the original 5-minute granularity to a 1-hour interval.
The detailed information is summarized in Table~\ref{dataset}. Moreover, we illustrate the node position of PEMS04 and PEMS08 with the latitude and longitude in Figure~\ref{map}. Following SSTBAN~\citep{sstban}, we normalize the data with the Z-score method and the mean and standard deviation are calculated from the training set.

In the long-term scenarios, the $T-T'$ is symmetric, and we set them as 24, 36, and 48 steps. In the extra-long-term scenarios, we set one scenario as 144-144 for half of the day prediction. For cases where the future prediction step $T'$ exceeds one day, $T$ is fixed to 288, and the future prediction step increases from 1 day to 7 days with a daily increment. This configuration considers a whole day as input, as it encompasses a complete daily cycle.

We slide the window to gain the input data $\mathbf{X}$ and ground truth $\mathbf{Y}$. We stride one step for each sample, and the length corresponds to the $T$ and $T'$.

\begin{table}[H]
        \caption{Summary of Datasets.}
        \centering
        \label{dataset}
        \begin{tabular}{@{}ccccc@{}}
        \toprule
        Dataset & \makecell{\# Time Steps} & \# Nodes    & Time Range        \\ \midrule
        PEMS04  & 16992      & 307                           & 2018/01 - 2018/02 \\
        PEMS08  & 17856      & 170                           & 2016/07 - 2016/08 \\
        Seattle Loop & 8760  & 323                           & 2015/01 - 2015/12 \\ \bottomrule
        \end{tabular}
\end{table}

\begin{table}[tbh]
\centering
\caption{Performance comparison on long-term traffic speed forecasting in Seattle Loop.}
\label{tab:seattleLong}
\begin{tabular}{@{}c|ccc|ccc|ccc@{}}
\toprule
Time step               & \multicolumn{3}{c}{24}              & \multicolumn{3}{c}{36}              & \multicolumn{3}{c}{48}              \\ Method
                     & \textit{RMSE} & \textit{MAE}  & \textit{MAPE}  & \textit{RMSE} & \textit{MAE}  & \textit{MAPE}  & \textit{RMSE} & \textit{MAE}  & \textit{MAPE}  \\ \midrule
HA      & 11.86 & 8.07 & 26.57 & 12.37 & 8.47 & 27.76 & 12.31 & 8.49 & 27.82 \\
VAR     & 9.56  & 6.21 & 19.94 & 9.96  & 6.44 & 21.28 & 10.28 & 6.69 & 22.24 \\
DCRNN   & 7.97  & 4.37 & 14.04 & 8.38  & 4.60 & 14.41 & 8.63  & 4.73 & 14.91 \\
GWNet   & 7.84  & 4.28 & 14.06 & 8.18  & 4.60 & 15.12 & 8.35  & 4.67 & 15.04 \\
GMAN    & 7.84  & 4.13 & 12.88 & 8.10  & 4.23 & 12.95 & 8.09  & 4.26 & 13.26 \\
AGCRN   & 7.83  & 4.27 & 13.53 & 8.31  & 4.66 & 14.76 & 8.60  & 4.82 & 15.62 \\
DMSTGCN & 7.59  & 4.08 & 13.51 & 7.98  & 4.31 & 14.31 & 8.20  & 4.49 & 14.86 \\
SSTBAN  & 7.72  & 4.05 & 12.69 & 7.83  & 4.11 & 12.44 & 7.88  & 4.12 & 12.25 \\
\textbf{Extralonger} & \textbf{7.43} & \textbf{4.04} & \textbf{12.48} & \textbf{7.51} & \textbf{4.05} & \textbf{11.96} & \textbf{7.68} & \textbf{4.11} & \textbf{12.04} \\ 
\bottomrule
\end{tabular}%
\end{table}

\begin{table}[tbh]
\centering
\caption{Performance comparison on extra-long-term traffic speed forecasting in Seattle Loop.}
\label{tab:seattleExtraLong}
\begin{tabular}{@{}c|ccc|ccc|ccc@{}}
\toprule
Method     & \multicolumn{3}{c}{HA} & \multicolumn{3}{c}{VAR} & \multicolumn{3}{c}{\textbf{Extralonger}}      \\ 
Time step & \textit{RMSE} & \textit{MAE} & \textit{MAPE} & \textit{RMSE} & \textit{MAE} & \textit{MAPE} & \textit{RMSE} & \textit{MAE} & \textit{MAPE} \\  \midrule 
144  & 11.95  & 8.30  & 27.48 & 11.17  & 7.33  & 24.75  & \textbf{8.07} & \textbf{4.33} & \textbf{12.64} \\
288  & 12.05  & 8.33  & 28.04 & 11.09  & 7.31  & 24.14  & \textbf{8.20} & \textbf{4.44} & \textbf{14.03} \\
576  & 11.95  & 8.14  & 27.88 & 11.30  & 7.43  & 24.83  & \textbf{8.15} & \textbf{4.59} & \textbf{14.46} \\
864  & 11.90  & 8.16  & 27.57 & 11.42  & 7.49  & 25.33  & \textbf{8.07} & \textbf{4.50} & \textbf{13.63} \\
1152 & 11.87  & 8.02  & 27.43 & 11.53  & 7.54  & 25.77  & \textbf{8.00} & \textbf{4.39} & \textbf{12.94} \\
1440 & 11.85  & 7.94  & 27.23 & 11.59  & 7.58  & 26.01  & \textbf{7.69} & \textbf{4.15} & \textbf{12.36} \\
1728 & 11.83  & 7.88  & 26.96 & 11.70  & 7.65  & 26.49  & \textbf{7.66} & \textbf{4.11} & \textbf{12.00} \\
2016 & 11.79  & 7.85  & 26.71 & 11.76  & 7.71  & 26.78  & \textbf{7.57} & \textbf{4.02} & \textbf{11.63} \\ \bottomrule
\end{tabular}%
\end{table}
\section{Experiments of Seattle Loop}\label{appendix:seattle}
 The results in the Seattle Loop dataset are presented in Table~\ref{tab:seattleLong} and Table~\ref{tab:seattleExtraLong} for long-term and extra-long-term scenarios, respectively. The best results are shown in bold. Our model \model~outperforms the baseline models for all metrics in all scenarios. Interestingly, in the extra-long-term scenarios, the prediction from \model becomes even better when the step length increases. We speculate that this improvement can be attributed to the increase in data volume, which allows the model to capture and learn long-range dependencies that may not be apparent in the shorter step length scenarios. The larger amount of data provides a richer context for the model to extract patterns and make more accurate predictions.

\section{Experiments of Error Bar}
\begin{figure}[H]
  \centering
  \subfigure{
    {\includegraphics[width=.31\textwidth]{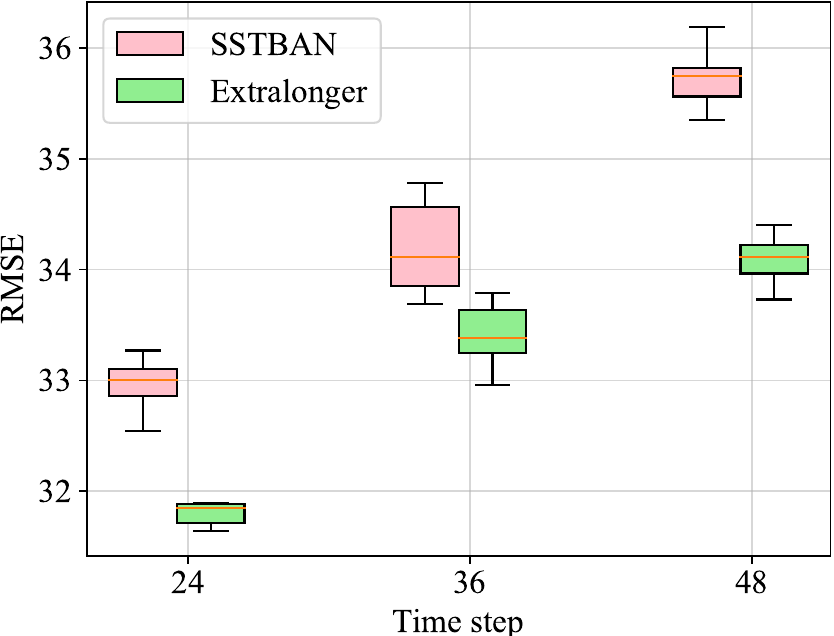}
}
  }  
  \subfigure{
    {\includegraphics[width=.31\textwidth]{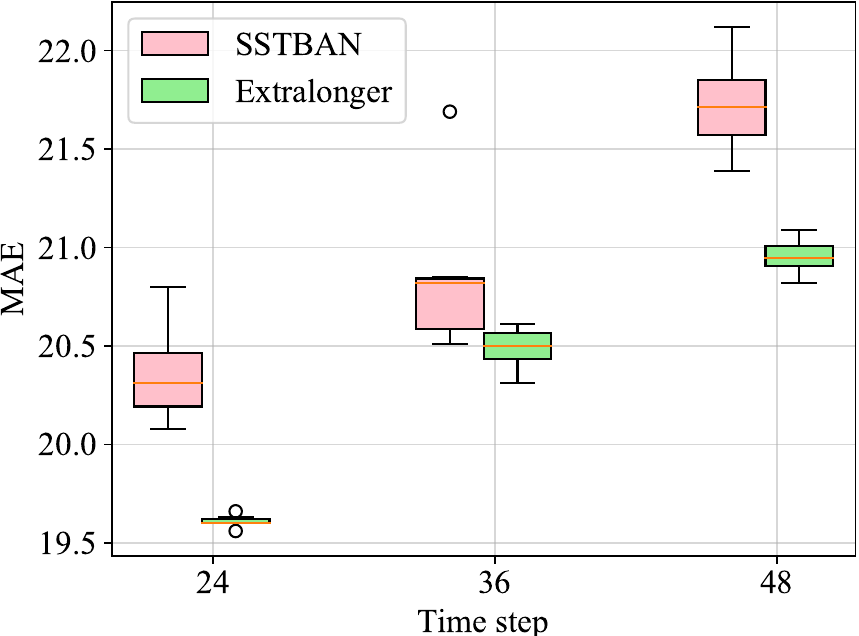}
}
  }
  \subfigure{
    {\includegraphics[width=.31\textwidth]{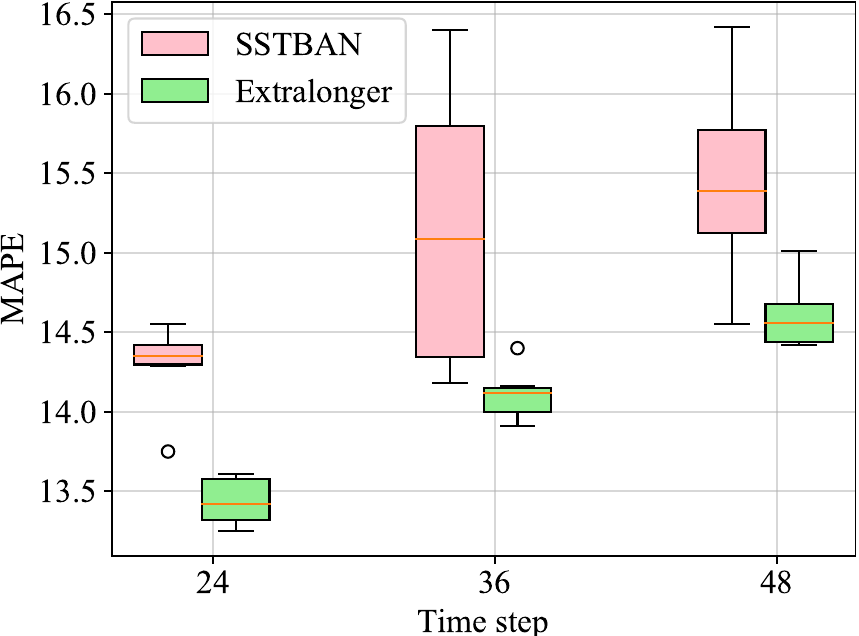}
}
  }
    \caption{Error bar on PEMS04 w.r.t RMSE (LEFT), MAE (MIDDLE) and MAPE (RIGHT).}
  \label{fig:errorbarpems04}
\end{figure}

\begin{figure}[H]
  \centering
  \subfigure{
    {\includegraphics[width=.31\textwidth]{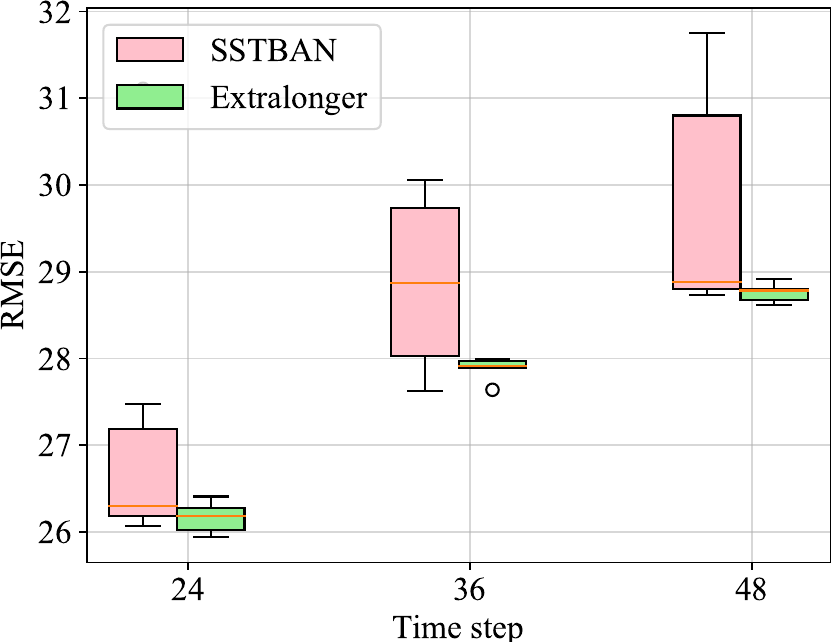}
}
  }  
  \subfigure{
    {\includegraphics[width=.31\textwidth]{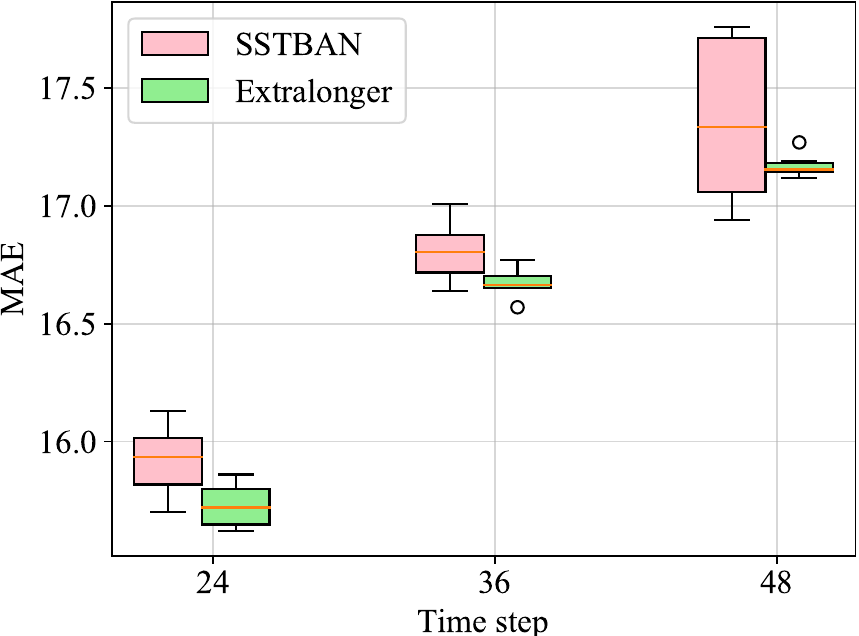}
}
  }
  \subfigure{
    {\includegraphics[width=.31\textwidth]{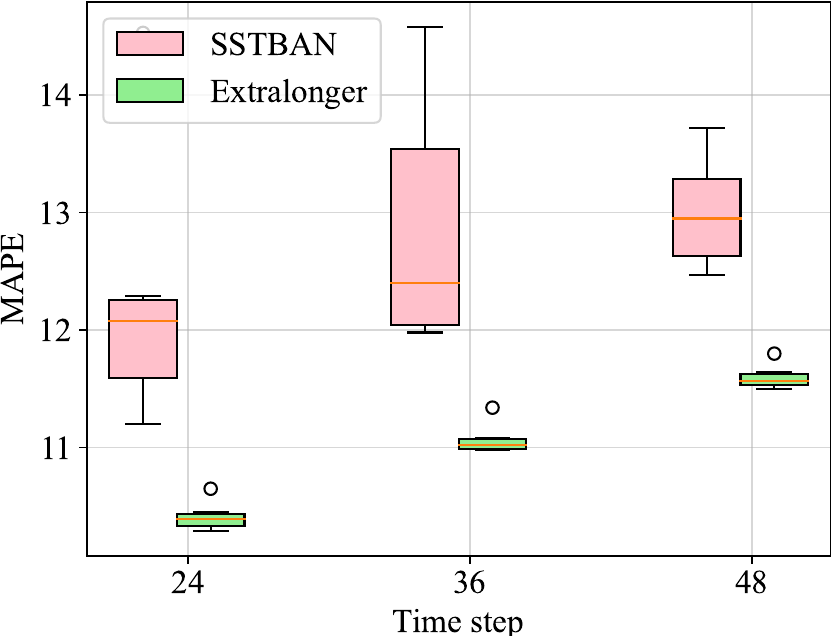}
}
  }
    \caption{Error bar on PEMS08 w.r.t RMSE (LEFT), MAE (MIDDLE) and MAPE (RIGHT).}
  \label{fig:errorbarpems08}
\end{figure}

We conducted error bar comparison experiments using box plots on PEMS04 and PEMS08 to compare our model with the prior best model, SSTBAN~\citep{sstban}. We analyzed six sets of results. As shown in Figure~\ref{fig:errorbarpems04} and Figure~\ref{fig:errorbarpems08}, our \model~consistently outperforms SSTBAN, as evidenced by smaller medians and shorter interquartile ranges. These findings indicate that our model demonstrates steady improvement over the prior best model and exhibits lower variability.

\section{Broader Impact}\label{appendix:broaderImpacts}

Our work could have a broader impact beyond traffic forecasting. It has the potential to significantly advance the field of transportation and potentially influence other general spatial-temporal tasks. Here are some key areas where our work could have a significant impact.

\paragraph{Transportation Field}
\begin{itemize}
    \item \textbf{Improved Traffic Management:} By enabling accurate extra-long-term traffic forecasting, \model~can empower traffic management authorities to proactively optimize traffic flow, reduce congestion, and improve overall transportation efficiency. 
    \item \textbf{Resource-Constrained Applications:} The remarkable reduction in resource consumption achieved by \model~unlocks the possibility of deploying traffic forecasting models on devices with limited computational power. This opens doors for real-time traffic prediction in resource-constrained environments, such as embedded systems in vehicles or edge computing devices.
    \end{itemize}

\paragraph{General Spatial-Temporal Tasks}\quad \\
The success of \model's unified spatial-temporal representation suggests a potential paradigm shift in approaching other tasks involving spatial and temporal dependencies. This \STRepresentation~could be explored and adapted to various domains requiring accurate long-horizon prediction based on interconnected spatial and temporal data, for example:
\begin{itemize}
    \item \textbf{Weather Forecasting:} Weather forecasting could benefit from a unified representation of spatial (e.g., atmospheric pressure, temperature) and temporal (e.g., historical weather patterns) data for improved long-term prediction.
    \item \textbf{Urban Planning:} Urban planning could utilize models enhanced by \STRepresentation~to forecast future resource demands (energy, water) based on spatial distribution and historical trends, enabling more sustainable infrastructure development.
\end{itemize}

Overall, \model~represents a significant advancement in traffic forecasting and paves the way for a more efficient approach to tackling various spatial-temporal modeling challenges.

\section{Limitations}\label{sec:limitation}
While \model~has successfully tackled the high-complexity issues inherent in classical representation, achieving a one-week prediction horizon, it still exhibits two certain limitations.

\model~effectively captures the dominant trend patterns in traffic flow data. However, \model's performance exhibits some degradation when encountering significant fluctuations, particularly during peak traffic hours (9:00-15:00). Real-world traffic data inherently exhibits these rapid variations, also known as pulse variations, which remain a challenge for current traffic forecasting models.  This limitation presents an exciting avenue for future research, aiming to develop more robust methods for capturing and predicting such short-term fluctuations.

Being fundamentally data-driven, our model relies solely on historical data for predictions and thus lacks the capability to respond to emergent traffic activities. This limitation may lead to potential misinterpretations within the ITS. Therefore, it is imperative to integrate additional information when planning and managing the traffic system.

\section{Future Work}   
It turns out that prediction errors tend to increase during periods of significant traffic flow fluctuations, particularly during peak hours (9:00-15:00). Real-world traffic data inherently exhibits pulse variations, which remain a challenging forecasting issue. We acknowledge this limitation and identify it as a promising avenue for future research.

Moreover, building upon the success of the Unified Spatial-Temporal Representation in \model, we aim to explore its applicability to a broader range of spatial-temporal tasks. This would demonstrate the generalization and efficiency of our proposed method, potentially establishing it as a versatile tool for various spatial-temporal tasks.


\newpage
\section*{NeurIPS Paper Checklist}

\begin{enumerate}

\item {\bf Claims}
    \item[] Question: Do the main claims made in the abstract and introduction accurately reflect the paper's contributions and scope?
    \item[] Answer: \answerYes{} 
    \item[] Justification: See the abstract and introduction part.

\item {\bf Limitations}
    \item[] Question: Does the paper discuss the limitations of the work performed by the authors?
    \item[] Answer: \answerYes{} 
    \item[] Justification: The forecasting error rises during 9:00-15:00, the peak traffic hours.

\item {\bf Theory Assumptions and Proofs}
    \item[] Question: For each theoretical result, does the paper provide the full set of assumptions and a complete (and correct) proof?
    \item[] Answer: \answerYes{} 
    \item[] Justification:  The complexity reduction theoretical analysis is based on some assumptions. We give the details in Section~\ref{sec:classicalComplexity} and Appendix~\ref{appendix:complexity}.

    \item {\bf Experimental Result Reproducibility}
    \item[] Question: Does the paper fully disclose all the information needed to reproduce the main experimental results of the paper to the extent that it affects the main claims and/or conclusions of the paper (regardless of whether the code and data are provided or not)?
    \item[] Answer: \answerYes{} 
    \item[] Justification: We introduce our method,~\model, step by step (Section~\ref{sec:model} and the detailed implementation is given in Section~\ref{sec:implementation}. The setup of experiments and the datasets are detailed in Appendix~\ref{appendix:datasetsAndSetup}.

\item {\bf Open access to data and code}
    \item[] Question: Does the paper provide open access to the data and code, with sufficient instructions to faithfully reproduce the main experimental results, as described in supplemental material?
    \item[] Answer: \answerYes{} 
    \item[] Justification: All the code and data are available in the supplementary material.

\item {\bf Experimental Setting/Details}
    \item[] Question: Does the paper specify all the training and test details (e.g., data splits, hyperparameters, how they were chosen, type of optimizer, etc.) necessary to understand the results?
    \item[] Answer: \answerYes{} 
    \item[] Justification: Training and test datasets partitioning details are given in Appendix~\ref{appendix:datasetsAndSetup}. The hyperparameters and optimizer are detailed in Section~\ref{sec:implementation}. 

\item {\bf Experiment Statistical Significance}
    \item[] Question: Does the paper report error bars suitably and correctly defined or other appropriate information about the statistical significance of the experiments?
    \item[] Answer: \answerYes{} 
    \item[] Justification: We report the RMSE, MAE and RMSE in long-term and extra-long-term scenarios (Section~\ref{sec:performance}), and we also gain a great reduction in training time, inference time and memory usage (Section~\ref{sec:resourceComparison}). 

\item {\bf Experiments Compute Resources}
    \item[] Question: For each experiment, does the paper provide sufficient information on the computer resources (type of compute workers, memory, time of execution) needed to reproduce the experiments?
    \item[] Answer: \answerYes{} 
    \item[] Justification: We conducted our experiment on one single NVIDIA 2080Ti GPU. The implementation details are given in Section~\ref{sec:implementation} and the experiment package dependency is in the supplementary material.
    
\item {\bf Code Of Ethics}
    \item[] Question: Does the research conducted in the paper conform, in every respect, with the NeurIPS Code of Ethics \url{https://neurips.cc/public/EthicsGuidelines}?
    \item[] Answer: \answerYes{} 
    \item[] Justification: We have checked the code of ethics, and our research is in line with the guidelines.

\item {\bf Broader Impacts}
    \item[] Question: Does the paper discuss both potential positive societal impacts and negative societal impacts of the work performed?
    \item[] Answer: \answerYes{} 
    \item[] Justification: See Appendix~\ref{appendix:broaderImpacts}
    
\item {\bf Safeguards}
    \item[] Question: Does the paper describe safeguards that have been put in place for responsible release of data or models that have a high risk for misuse (e.g., pretrained language models, image generators, or scraped datasets)?
    \item[] Answer: \answerNA{} 
    \item[] Justification: Our research is for traffic forecasting, which does not have the misuse risk mentioned in the question.

\item {\bf Licenses for existing assets}
    \item[] Question: Are the creators or original owners of assets (e.g., code, data, models), used in the paper, properly credited and are the license and terms of use explicitly mentioned and properly respected?
    \item[] Answer: \answerYes{} 
    \item[] Justification: All the code we use is from open-source libraries. The datasets are the public dataset. And we cited them properly.
    \item[] Guidelines:
    \begin{itemize}
        \item The answer NA means that the paper does not use existing assets.
        \item The authors should citep the original paper that produced the code package or dataset.
        \item The authors should state which version of the asset is used and, if possible, include a URL.
        \item The name of the license (e.g., CC-BY 4.0) should be included for each asset.
        \item For scraped data from a particular source (e.g., website), the copyright and terms of service of that source should be provided.
        \item If assets are released, the license, copyright information, and terms of use in the package should be provided. For popular datasets, \url{paperswithcode.com/datasets} has curated licenses for some datasets. Their licensing guide can help determine the license of a dataset.
        \item For existing datasets that are re-packaged, both the original license and the license of the derived asset (if it has changed) should be provided.
        \item If this information is not available online, the authors are encouraged to reach out to the asset's creators.
    \end{itemize}

\item {\bf New Assets}
    \item[] Question: Are new assets introduced in the paper well documented and is the documentation provided alongside the assets?
    \item[] Answer: \answerYes{} 
    \item[] Justification: The code is available in the supplementary material and we give a simple README.md file to introduce the launch commands.

\item {\bf Crowdsourcing and Research with Human Subjects}
    \item[] Question: For crowdsourcing experiments and research with human subjects, does the paper include the full text of instructions given to participants and screenshots, if applicable, as well as details about compensation (if any)? 
    \item[] Answer: \answerNA{} 
    \item[] Justification: Our research does not involve crowdsourcing or research with human subjects.

\item {\bf Institutional Review Board (IRB) Approvals or Equivalent for Research with Human Subjects}
    \item[] Question: Does the paper describe potential risks incurred by study participants, whether such risks were disclosed to the subjects, and whether Institutional Review Board (IRB) approvals (or an equivalent approval/review based on the requirements of your country or institution) were obtained?
    \item[] Answer: \answerNA{} 
    \item[] Justification: Our research does not involve crowdsourcing or research with human subjects.
    
\end{enumerate}

\end{document}